\documentclass[11pt]{article}

\usepackage[final]{acl}

\usepackage{times}
\usepackage{latexsym}
\usepackage{makecell}
\usepackage[T1]{fontenc}

\usepackage[utf8]{inputenc}

\usepackage{microtype}

\usepackage{inconsolata}

\usepackage{orcidlink}
\usepackage{graphicx}
\usepackage{amsmath}
\usepackage{multirow}
\usepackage{CJKutf8}
\usepackage{times}
\usepackage{soul}
\usepackage{url}
\usepackage{amssymb}
\usepackage{amsthm}
\usepackage{xcolor} 
\usepackage{booktabs}
\usepackage{algorithm}
\usepackage{algorithmic}
\usepackage[switch]{lineno}
\usepackage{xcolor}
\usepackage{siunitx}
\usepackage[normalem]{ulem} 
\newcommand{\xmark}{\textsuperscript{$\divideontimes$}}
\newcommand{\repro}{\rlap{\textsuperscript{\(\divideontimes\)}}}
\newcommand{\secref}[1]{\S\kern-0.05em\ref{#1}}

\newcommand{\ModelName}{\text{BioELX}}
\newcommand{\Mention}{m}
\newcommand{\Text}{T}
\newcommand{\Entity}{e}

\newcommand{\Encoder}{g}
\newcommand{\Input}{\Phi}
\newcommand{\Ranker}{R_\theta}
\newcommand{\Alias}{a}
\newcommand{\Qid}{q}
\newcommand{\Cui}{c}
\newcommand{\Language}{\ell}
\newcommand{\SapBERTmulti}{SapBERT$_{\mbox{multi}}$}

\newcommand{\KBase}{\mathcal{K}}
\newcommand{\Ppairs}{\mathcal{P}}
\newcommand{\Npairs}{\mathcal{N}}
\newcommand{\Candidate}{C}
\newcommand{\Mentions}{M}

\newcommand{\Aliases}{\mathcal{A}}

\title{BioELX: Context-Aware Cross-lingual Biomedical Entity Linking without Task-Specific Supervision}

\author{
 \textbf{Yi Wang\orcidlink{0000-0001-6086-807X}\textsuperscript{1}},
 \textbf{Corina Dima\orcidlink{0000-0001-7409-4992}\textsuperscript{1}},
 \textbf{Liangyu Zhong\orcidlink{0009-0000-4286-9095}\textsuperscript{2}},
 \textbf{Steffen Staab\orcidlink{0000-0002-0780-4154}\textsuperscript{1,3}},
\\
\\
 \textsuperscript{1}Institute for AI, University of Stuttgart,
 \\
 \textsuperscript{2}Technical University of Berlin,
 \textsuperscript{3}University of Southampton,
\\
 \small{\textbf{Correspondence:} 
   \href{mailto:email@domain}{yi.wang@ki.uni-stuttgart.de}
 }
}

\begin{document}
\maketitle
\begin{abstract}
Cross-lingual biomedical entity linking (BEL) maps mentions in any language to unique identifiers in a biomedical knowledge base, supporting clinical and biomedical NLP applications. We identify two issues affecting current systems. First, the UMLS \cite{DBLP:journals/nar/Bodenreider04} aliases used to train cross-lingual BEL retrievers are heavily skewed toward English, so retrievers generalize poorly to non-English mentions. Second, although context is often necessary for disambiguation, naively injecting context into retrievers trained only to align aliases severely degrades retrieval. We propose \ModelName{}, a retrieve--rerank framework that addresses both issues. For retrieval, we continue training \SapBERTmulti{} \cite{DBLP:conf/acl/0001VKC20} using \textit{Wikidata-derived cross-lingual alias supervision}, forming shared concept neighborhoods across languages. For reranking, we adapt pretrained LLM rerankers to entity linking through \textit{mention-anchored prompting}, which marks the target mention so that rerankers score candidates with respect to the intended mention rather than other salient tokens in the context. Experiments show that \ModelName{} achieves new state-of-the-art results on four cross-lingual BEL benchmarks, improving Recall@1 by 4.9 to 18.2 percentage points over prior best results, \textit{without any task-specific BEL annotations}. Our code and resources are available at \url{https://github.com/AI4MedCode/BioELX}.
\end{abstract}

\section{Introduction}
\label{sec:intro}

Entity linking (EL) maps an ambiguous text span called a \textit{mention} in a given text to the corresponding \textit{entity} in a \textit{knowledge base (KB)}, which stores entities as \textit{unique IDs} associated with names, aliases, and optional metadata. Biomedical entity linking (BEL) is a domain-specific variant of EL, where mentions refer to biomedical entities such as diseases, drugs, symptoms, etc. Cross-lingual biomedical entity linking extends BEL to the multilingual setting, where mentions may appear in any language but must be linked to the same KB entities. 
In the example sentence ``\textit{Une réduction du nombre de globules rouges peut entraîner des symptômes tels que fatigue ou \textbf{essoufflement}.''}\footnote{Translation: A reduction in the number of red blood cells can lead to symptoms such as fatigue or shortness of breath.} in Fig.~\ref{fig:pipeline},
the mention \textit{essoufflement} should be linked to CUI C0013404, which has multilingual aliases such as
\textit{dyspnea} and \textit{shortness of breath} in English, and \textit{Sesak napas} in Indonesian. \textit{CUI} stands for \textit{concept unique identifier} and denotes IDs in the UMLS~\cite{DBLP:journals/nar/Bodenreider04}, a comprehensive biomedical KB. However, KBs do not always contain an alias in the mention language: here, CUI C0013404 has no French alias, so the linker must connect the French mention to the correct concept through cross-lingual evidence. 

Monolingual EL systems commonly adopt a two-stage design: \textit{candidate retrieval} followed by \textit{candidate reranking}~\cite{DBLP:conf/emnlp/WuPJRZ20,DBLP:conf/iclr/CaoI0P21,DBLP:conf/acl/OrlandoCBN24}. This design balances efficiency and accuracy. The lightweight \textit{retriever} efficiently selects a top-$k$ candidate set from a KB over a precomputed entity embedding index, aiming for broad candidate coverage that includes the gold entity. A more expressive \textit{reranker} then leverages richer information, such as mention context, to choose the final prediction. Reranking is more accurate but also more computationally intensive, because many rerankers jointly score each mention--candidate pair, requiring a separate forward pass per pair and limiting the use of pre-computed entity representations \cite{DBLP:conf/emnlp/WuPJRZ20}. Moreover, training such rerankers typically requires a substantial number of task-specific EL annotations.

In cross-lingual BEL, many existing systems are built on \SapBERTmulti{}~\cite{DBLP:conf/acl/0001VKC20} as the retrieval component. Some extend or fine-tune it for specific languages and domains~\cite{sakhovskiy-etal-2024-biomedical,mustafa-etal-2024-leveraging,Fierens_umls_trans}; others add a second-stage disambiguation model trained on task-specific BEL annotations~\cite{DBLP:conf/emnlp/ZhuQCMYX23,DBLP:journals/corr/abs-2310-11275,DBLP:conf/iccS/GallegoLSKV24}, which are scarce for many languages.

We identify two issues with current cross-lingual BEL models. First, the retriever \SapBERTmulti{} generalizes poorly to non-English mentions because its training resources provide limited multilingual evidence. This is apparent in the alias distribution used to train \SapBERTmulti{}, where $69.6\%$ of the aliases are written in English, while the remaining $30.4\%$ are spread across 22 other languages~\cite{DBLP:conf/acl/0001VKC20}. This imbalance causes the learned representations to favor English surface forms, so non-English mentions are systematically drawn toward lexically similar but semantically incorrect entities. As shown in Fig.~\ref{fig:sapfail}, the French mention \textit{essou\textbf{ffle}ment} should be linked to the CUI C0013404 (\textit{Dyspnea}), but \SapBERTmulti{} retrieves instead the lexically similar but semantically different concept \textit{E\textbf{ffle}urage} (`light massage', CUI C0203907), likely because both strings share the rare substring \textbf{-ffle}.

This motivates our first hypothesis (H1): broader cross-lingual alias coverage should reduce the retriever's reliance on string overlap and enable non-English mentions to be linked through cross-lingual evidence.  
We test this hypothesis with \textit{Wikidata-derived cross-lingual alias supervision}, which provides broader multilingual alias coverage for retriever training by mapping Wikidata aliases to biomedical concepts. \secref{sec:method-alias} details the alias extraction, mapping, and training procedure.

A second issue is that contextual disambiguation is difficult to incorporate without sacrificing retrieval quality or relying on task-specific supervision.
\textit{Alias-based retrievers} such as \SapBERTmulti{} are trained to match aliases of the same concept, not to encode a mention with its context. Applying them to inputs that combine the mention with its context can distort the learned alias representation, shifting it away from the underlying concept and severely degrading retrieval. This effect is confirmed by our diagnostic experiment in Appendix~\ref{app:context_retrieval}: adding context reduces \SapBERTmulti{} Recall@1 by more than 50 percentage points on both tested benchmarks. 
A standard approach to incorporate context is to use a dedicated context-aware reranker, but training such a reranker requires high-quality expert annotations that link mention spans to unambiguous entity identifiers. Creating such data is costly and such datasets are scarce, particularly for low-resource languages. 

Pretrained LLM rerankers offer a training-free alternative because they are trained to compare query--document relevance, enabling context-aware candidate comparison without BEL-specific reranker training. We use rerankers rather than generative LLMs because entity generation can produce hallucinated or out-of-KB strings~\cite{DBLP:journals/csur/JiLFYSXIBMF23}, whereas rerankers score candidates directly. 

However, applying off-the-shelf LLM rerankers to EL introduces a task mismatch. These models are trained for general query--document relevance, not for entity disambiguation at a specific mention span. They may therefore focus on salient contextual entities instead of the target mention, which we call the \textit{focus-mismatch problem}. For example, when reranking candidates for the mention \textit{essoufflement} in Fig.~\ref{fig:temp}, a reranker may attend to \textit{globules rouges} and incorrectly prefer \textit{Red Blood Cells} over the correct entity \textit{Dyspnea}. This leads to our second hypothesis (H2): pretrained LLM rerankers can provide context-aware reranking without supervised BEL annotations, but must be adapted to score candidates with respect to the target mention rather than the whole context. We instantiate this adaptation with \textit{mention-anchored prompting}, which marks the target mention in the input so that LLM rerankers score candidates against it. \secref{sec:llm} details this adaptation.



Our contributions are: \textbf{1)} \ModelName{}, a cross-lingual biomedical entity linking framework that can be applied to new benchmarks without training on any BEL-annotated data. \ModelName{} achieves new state-of-the-art results on four cross-lingual benchmarks, improving the previous best by \textbf{+18.2} R@1 on XL-BEL~\cite{DBLP:conf/acl/0001VKC20}, \textbf{+4.9} on EMEA, \textbf{+8.9} on Patent~\cite{10.1093/jamia/ocv037}, and \textbf{+7.8} on WikiMed-DE~\cite{DBLP:conf/wikidata/WangDS23}. Gains are especially pronounced for low-resource languages (Turkish \textbf{+18.2}, Korean \textbf{+21.1}, and Thai \textbf{+27.1} on XL-BEL); \textbf{2)} a Wikidata-derived cross-lingual resource for biomedical retrieval, providing broad multilingual concept-level supervision and \textbf{3)}
a systematic study of how to adapt pretrained LLM rerankers to EL, that identifies mention-anchored prompting as a simple, training-free recipe for context-aware reranking.
The architecture of \ModelName{} is shown in Fig.~\ref{fig:pipeline}. 
Following prior work~\cite{DBLP:conf/emnlp/WuPJRZ20,DBLP:conf/naacl/0003AMM22,sakhovskiy-etal-2024-biomedical}, we assume that mention spans are given and focus on linking each mention to a KB entity (see \secref{sec:pf} for details).

\section{Related Work}
\label{sec:rw}

\subsection{Biomedical Entity Linking}
BEL systems commonly follow a two-stage retrieve-then-rerank pipeline. Across both stages, prior work faces a recurring trade-off: methods that exploit mention context for disambiguation require task-specific BEL annotations, while annotation-free methods rely on KB aliases alone and ignore context.

\paragraph{Retrieval.} Two paradigms dominate: the first trains retrievers on KB aliases with a contrastive objective, mapping different surface forms of the same concept to nearby embeddings~\cite{DBLP:conf/acl/SungJLK20,DBLP:conf/naacl/LiuSMBC21,DBLP:journals/jbi/YuanZSLWY22,10.1093/bioinformatics/btae474}. It uses only KB aliases as supervision and does not condition on mention context. The second paradigm trains bi-encoders on annotated mention--entity pairs \cite{DBLP:conf/emnlp/WuPJRZ20}, encoding the mention in its context and the entity into a shared space~\cite{DBLP:conf/naacl/0003AMM22,DBLP:conf/aaai/XuCH23,lin-etal-2024-biomedical}; this exploits context but requires task-specific BEL annotations.

\paragraph{Reranking.} Rerankers, in contrast, all incorporate mention context and differ mainly in how candidates are compared. Many score each mention--candidate pair independently, jointly encoding the mention context and one candidate entity to assign a relevance score~\cite{DBLP:conf/naacl/0003AMM22,sanz-cruzado-lever-2025-accelerating}. Others place the mention context and several candidates in a shared input so they can be compared directly: Prompt-BioEL~\cite{DBLP:conf/aaai/XuCH23} and BioELQA~\cite{lin-etal-2024-biomedical} cast linking as a multiple-choice template, scored by predicting which candidate fills a masked slot in the template for Prompt-BioEL and using a generator to output the chosen candidate for BioELQA. In either way, training relies on task-specific BEL annotations.

\paragraph{Cross-lingual BEL.} \citet{DBLP:conf/acl/0001VKC20} introduced the XL-BEL benchmark and \SapBERTmulti{}, a multilingual alias-based retriever trained contrastively on UMLS aliases, which has become the standard backbone for cross-lingual BEL. Many subsequent systems build directly on it. On the retrieval side, \citet{mustafa-etal-2024-leveraging} and \citet{Fierens_umls_trans} fine-tune \SapBERTmulti{} for German and French, and \citet{sakhovskiy-etal-2024-biomedical} add KB concept relations as a training signal. A complementary line of work adds a disambiguation stage after retrieval: Con2GEN~\cite{DBLP:conf/emnlp/ZhuQCMYX23} trains a generative model to produce the target entity name, and ClinLinker~\cite{DBLP:conf/iccS/GallegoLSKV24} trains a cross-encoder over mention context and candidate. Yet, retriever-side approaches still inherit the English-skewed alias supervision of UMLS and are typically tuned for a single language, while a disambiguation stage depends on BEL annotations that are scarce for most languages. \ModelName{} targets both gaps: it broadens alias supervision with cross-lingual Wikidata aliases and adds context-aware reranking without BEL annotations (\secref{sec:methodology}).

\subsection{LLMs in Entity Linking}
\label{sec:elllm}
Prior work has used LLMs for EL in two main ways. The first treats the LLM as an auxiliary component of an existing EL pipeline, e.g.,  rewriting or augmenting mention context~\cite{DBLP:conf/cikm/XinQ0ZZ00L25,vollmers-etal-2025-contextual} or simplifying mentions to better match KB aliases~\cite{DBLP:journals/biodb/BorchertLS24}. The second reformulates EL as a generation task, where the LLM directly generates the target entity name or identifier, optionally constrained to a set of retrieved candidates~\cite{DBLP:conf/emnlp/XiaoGWZSJ23,DBLP:conf/coling/Ding0W24,DBLP:conf/sigir/Xie0HN0024}. However, given the diversity of entity aliases, generation-based linking is prone to instruction-following failures and hallucinated or out-of-KB outputs~\cite{DBLP:journals/csur/JiLFYSXIBMF23}. 

A third direction, largely unexplored for EL, is to use off-the-shelf LLM rerankers - models trained on large-scale query-document relevance data and widely used as rerankers in information retrieval \cite{DBLP:journals/corr/abs-2402-03216, DBLP:journals/corr/abs-2509-25085, DBLP:journals/corr/abs-2506-05176}. Unlike generation models, rerankers score retrieved candidates and therefore cannot produce out-of-KB outputs; unlike pipeline-augmentation methods, they directly compare the mention with candidate entities. They are, however, trained for general query-document relevance rather than for mention-level disambiguation. \ModelName{} adapts pretrained LLM rerankers to cross-lingual BEL without task-specific BEL annotations (\secref{sec:llm}).

\section{Methodology}
\label{sec:methodology}
\subsection{Problem Formulation}
\label{sec:pf}
Let $\Text=(t_1, \dots, t_N)$ be a biomedical text of $N$ tokens. We assume that a set of mentions $\Mentions=\{\Mention_1,\ldots,\Mention_{|\Mentions|}\}$, has already been identified in $\Text$, where each mention $\Mention_i=(t_{\Mention_{i,1}},\ldots,t_{\Mention_{i,n_i}})$ is a contiguous span of $n_i$ tokens in $\Text$ that refers to a biomedical concept. The target knowledge base $\KBase=\{\Entity_1,\ldots,\Entity_{|\KBase|}\}$ contains $|\KBase|$ entities; each entity $\Entity$ has a unique identifier, an alias set $\Aliases(\Entity)=\{\Alias_1,\ldots,\Alias_{|\Aliases(\Entity)|}\}$ of surface forms in one or more languages, and optional metadata such as semantic types and textual descriptions. 

Cross-lingual biomedical entity linking maps each mention to its corresponding KB entity. A mention may be written in any language, and that language need not be covered by the aliases of its gold entity in $\KBase$; this mismatch between the mention language and the KB is the defining difficulty of the cross-lingual setting. We assume each mention links to exactly one entity in $\KBase$. Formally, the task is to determine a linking function $f$ such that $f(\Mention,\Text,\KBase)=\Entity$ for each mention $\Mention \in \Mentions$ and its gold entity $\Entity \in \KBase$.

\subsection{$\ModelName$ Framework}
\label{sec:framework}

\begin{figure}[h]
  \includegraphics[width=\columnwidth]{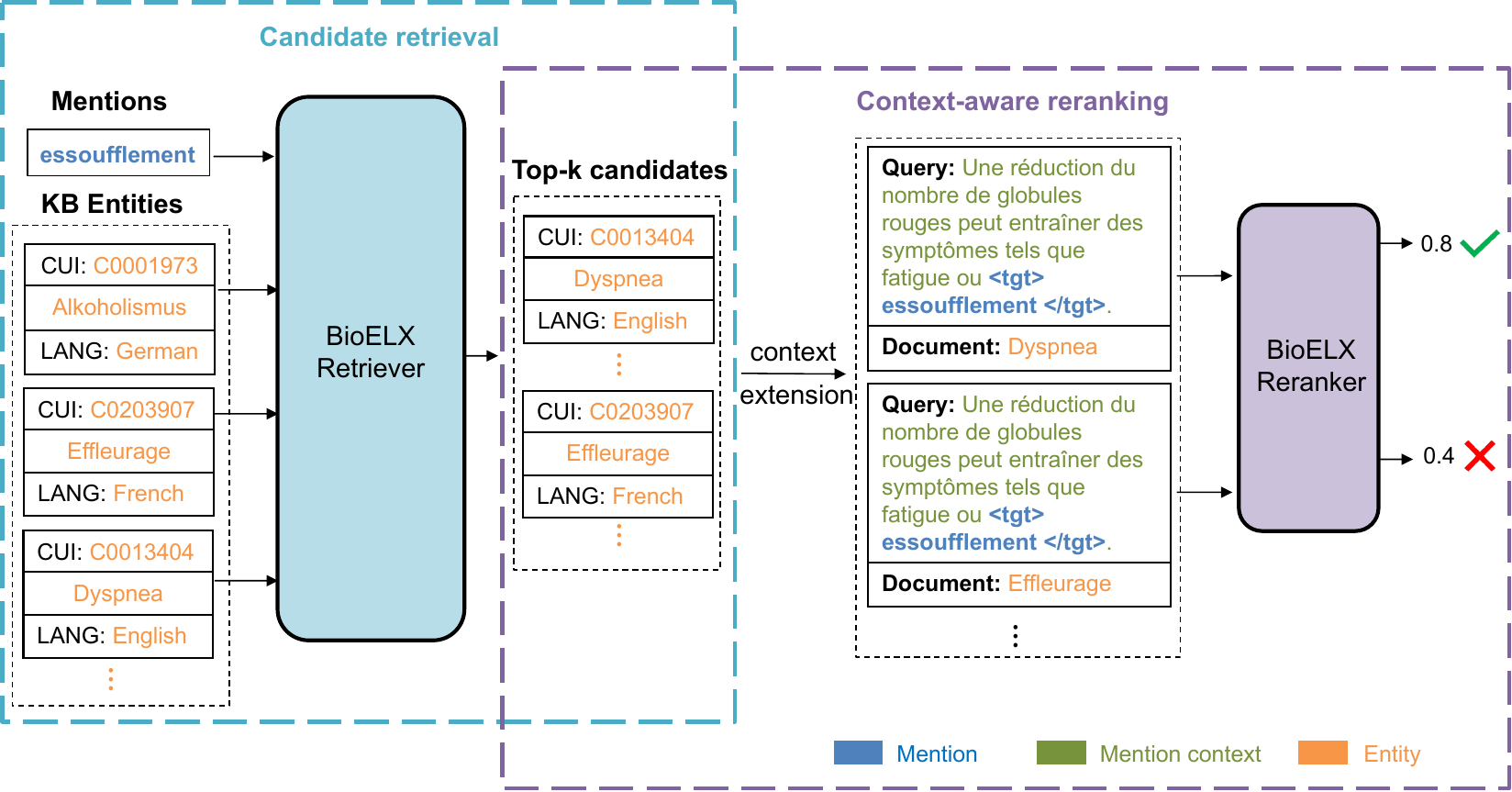}
  \caption{The two-stage \ModelName{} pipeline, illustrated for a single mention. The retriever embeds the mention and multilingual entity aliases into a shared space, and returns the top-$k$ aliases with associated entities by cosine similarity. For each candidate alias, the reranker scores a \textit{mention-anchored query} --- the mention and its surrounding context --- against a document built from that alias, and selects the entity associated with the highest-scoring alias. Larger version in Figure~\ref{fig:pipeline_large}.}
  \label{fig:pipeline}
\end{figure}

\noindent Figure~\ref{fig:pipeline} shows the overall pipeline of \ModelName{}, which comprises the \ModelName{} Retriever for cross-lingual candidate retrieval and the \ModelName{} Reranker for context-aware reranking.

\paragraph{Candidate retrieval.}
A single encoder $\Encoder(\cdot)$ embeds both mentions and aliases in a shared space. For efficiency, we embed every alias of every entity $\Entity \in \KBase$ offline and index the resulting vectors; at inference time, we embed only the input mention $\Mention$. Each alias is scored by its cosine similarity to the mention, 

\begin{equation}
 s_{\text{ret}}(\Mention,\Alias)=\cos(\Encoder(\Mention),\Encoder(\Alias))   
\end{equation}

\noindent and we take the $k$ highest-scoring aliases as the candidate set $\Candidate_k(\Mention)$, each paired with its associated entity.

\begin{figure}[h]
  \includegraphics[width=\columnwidth]{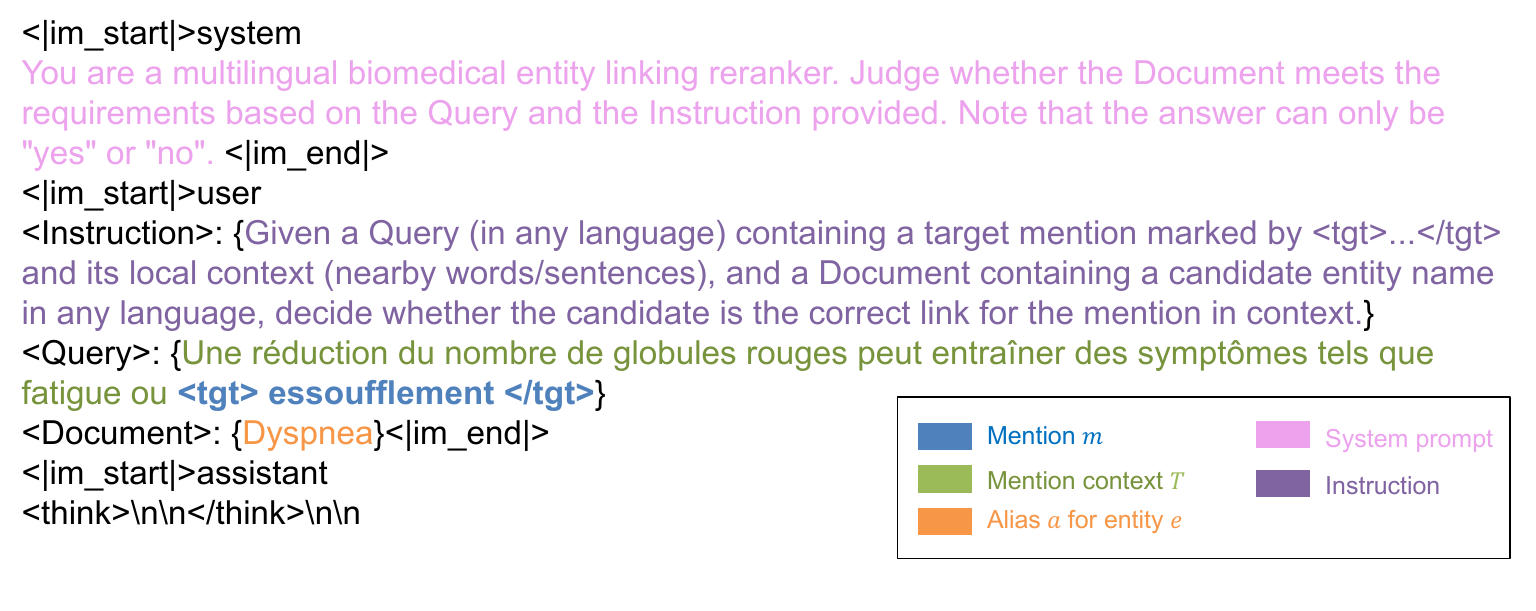}
  \caption{An example of the \ModelName{} Reranker input, instantiated with the \textsc{Qwen}3-Reranker template (black text). The \textbf{query} contains the biomedical text $\Text$ with the target mention $\Mention$ marked by \texttt{<tgt>}...\texttt{</tgt>}, and the \textbf{document} contains an alias $\Alias$ of the candidate entity $\Entity$, so the reranker scores each mention-candidate pair. Our ``document'' is intentionally short: many KB entities lack rich textual metadata, so we provide only the entity alias. The same mention-anchored input construction can be incorporated into the native scoring template of other LLMs. See larger version in Fig.~\ref{fig:temp_large}.}
  
  \label{fig:temp}
\end{figure}

\paragraph{Context-aware reranking.}
Each candidate alias $\Alias \in C_k(\Mention)$ is then rescored in context. We construct a query--document input 
$\Input\big(Q(\Mention,\Text),D(\Alias)\big)$, where the query $Q(\Mention,\Text)$ is built from the mention and its surrounding context, and the document $D(\Alias)$ contains an alias in $C_k(\Mention)$ (see Figure~\ref{fig:temp}). A pretrained LLM reranker $\Ranker$  assigns a relevance score
\begin{equation}
s_{\mathrm{rank}}(\Mention,\Alias \mid \Text)
=
\Ranker(\!\Input\big(Q(\Mention,\Text),D(\Alias)\big))
\in \mathbb{R}.
\end{equation}
and \ModelName{} predicts the highest-scoring alias,
\begin{equation}
\Alias_{\text{final}}
=
\arg\max_{\Alias\in C_k(\Mention)}
s_{\mathrm{rank}}(\Mention,\Alias \mid \Text).
\end{equation}
\noindent
and outputs its associated entity. The \ModelName{} Reranker is independent of the underlying model architecture and scoring interface. Appendix~\ref{app:scoring-variants} presents additional scoring variants evaluated in our experiments, including confidence-based scoring for \textsc{Qwen}3-Reranker~\cite{DBLP:journals/corr/abs-2506-05176}, listwise scoring for Jina-Reranker~\cite{DBLP:journals/corr/abs-2509-25085}, and logit-based scoring for BGE-Reranker~\cite{DBLP:journals/corr/abs-2402-03216}. \secref{sec:llm} specifies how \ModelName{} constructs the mention-anchored query $Q(\Mention,\Text)$ and candidate document $D(\Alias)$ for LLM rerankers.

\subsection{Wikidata-derived Cross-lingual Retrieval}
\label{sec:method-alias}
\begin{figure}[h]
  \includegraphics[width=\columnwidth]{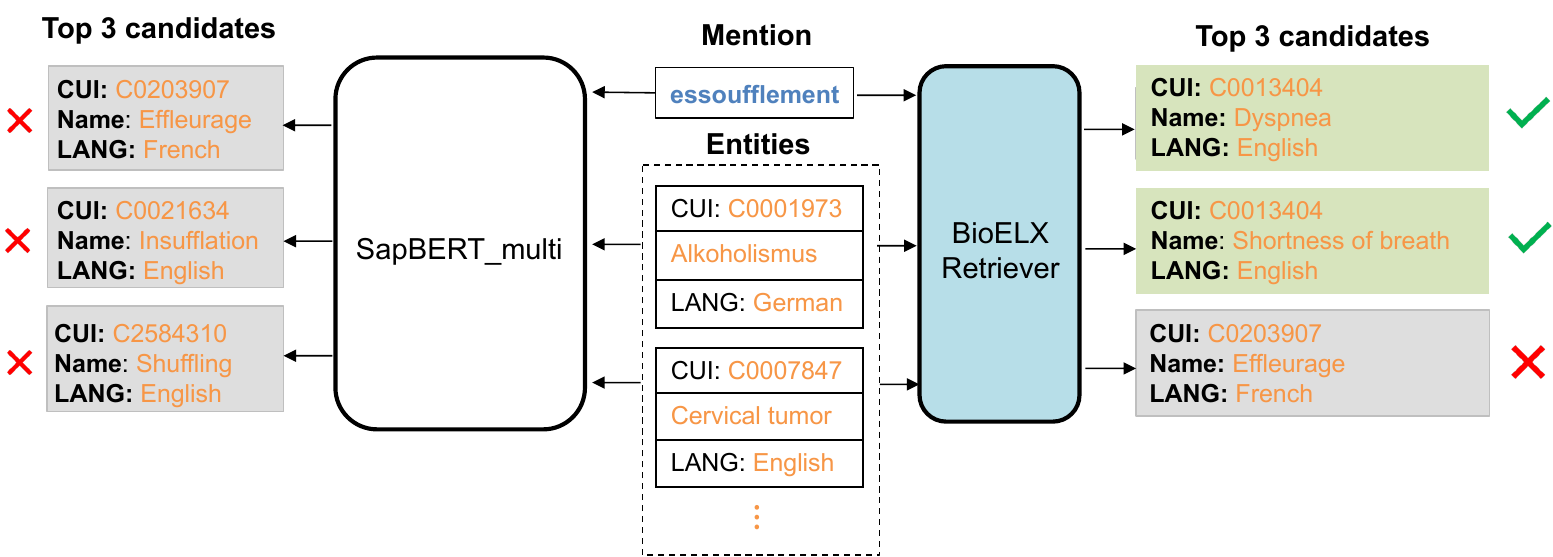}
  \caption{An example of retrieval. For the French mention \textit{essoufflement}, \SapBERTmulti{} retrieves top candidates based on surface similarity but misses the correct concept. Our retriever, enhanced with Wikidata-derived multilingual aliases, retrieves candidates that include the correct CUI (Dyspnea) among the top results. See larger version in Fig.~\ref{fig:sapfail_large}.}
  \label{fig:sapfail}
\end{figure}

To address the issue of \SapBERTmulti{} discussed in Section~\ref{sec:intro}, we construct \textit{Wikidata-derived cross-lingual alias supervision}: instead of relying only on UMLS aliases, we use Wikidata-derived multilingual aliases to create cross-lingual concept-level alias groups for retriever training. The intuition is that aliases of the same biomedical concept in different languages can serve as cross-lingual bridges in the embedding space. By pulling these aliases together during contrastive training, the retriever learns shared multilingual concept neighborhoods, so an unseen non-English mention can be drawn toward the correct concept rather than toward a lexically similar but semantically unrelated entity. We instantiate this idea by extracting multilingual aliases from Wikidata~\cite{DBLP:journals/cacm/VrandecicK14}, mapping them to UMLS, and continuing contrastive training from \SapBERTmulti{}. Figure~\ref{fig:sapfail} provides an illustrative example. For the French mention \textit{essoufflement}, Wikidata-derived multilingual aliases provide broader cross-lingual evidence that helps place the correct entity closer to this mention in the embedding space, so it appears among the retrieved top-$k$ candidates.


\paragraph{Constructing cross-lingual alias supervision from Wikidata.}
To provide multilingual alignment supervision beyond the English-skewed aliases in UMLS, we derive cross-lingual aliases from a recent Wikidata dump\footnote{\label{pageprops} \url{https://dumps.wikimedia.org/}, version 2025-12-01.} and map them to UMLS CUIs through Wikidata identifiers (QIDs). Here, an \emph{alias} denotes a language-specific surface form associated with a QID. 

Concretely, we parse the Wikidata \texttt{item\_per\_site} dump to extract triples of the form $(\Qid,\Alias,\Language)$, where $\Qid$ is a QID, $\Alias$ is a language-specific alias, and $\Language$ denotes the corresponding language key. A single QID can have aliases in multiple languages. For example, \texttt{(3358860, `Ovalocyte', frwiki)} indicates that the QID 3358860 has the French alias \texttt{`Ovalocyte'}. We collect all aliases in all languages for a given QID. 

We then map each QID to a UMLS CUI by querying the Wikidata SPARQL endpoint\footnote{\url{https://query.wikidata.org/}} for items annotated with the property \texttt{P2892, UMLS CUI}\footnote{ \url{https://www.wikidata.org/wiki/Property:P2892}}. Attaching this mapping to the extracted triples yields four-tuples $(\Qid,\Alias,\Language,\Cui)$, where $\Cui$ is the associated CUI, e.g., \texttt{(3358860, `Ovalocyte', frwiki, C0221280)}. We retain only tuples with a valid CUI, which serves as a filtering step to restrict the collection to biomedical aliases. We discuss adaptations of this filtering step to other biomedical resources in Appendix~\ref{app:adapt_wiki_ontology}.

To ensure that our retrieval gains reflect zero-shot generalization to unseen mentions rather than memorization of evaluation surface forms, we discard any tuple whose alias string exactly matches a mention string in our evaluation benchmarks.


We obtain $3,823,814$ aliases across $597$ languages for $537,848$ unique UMLS CUIs. English accounts for only $6.3\%$ of these aliases, while the nine non-English languages covered by XL-BEL together account for $11.3\%$. This sharply contrasts with the UMLS-derived aliases used to train \SapBERTmulti{}, where English accounts for $69.6\%$~\cite{DBLP:conf/acl/0001VKC20}. Detailed statistics are provided in Appendix~\ref{app:data-stat}.

\paragraph{Contrastive Training on Wikidata-Derived Aliases.}
Starting from \SapBERTmulti{}\footnote{SapBERT-UMLS-2020AB-all-lang-from-XLMR-large}, we continue training the encoder $\Encoder(\cdot)$ on Wikidata-derived cross-lingual aliases with the same contrastive learning objective, expanding multilingual alias coverage beyond UMLS and mitigating the English bias in alias supervision.
Given a mini-batch $\mathcal{B}=\{(x_i,y_i)\}_{i=1}^{N}$, where $x_i$ is an alias string and $y_i$ is its CUI label, we define for the \textit{anchor alias} $x_i$ the in-batch positive and negative sets as
\begin{align}
\Ppairs(x_i)&=\{x_j \mid (x_j,y_j)\in\mathcal{B},\ x_j\neq x_i,\ y_j=y_i\}\\
\Npairs(x_i)&=\{x_j \mid (x_j,y_j)\in\mathcal{B},\ x_j\neq x_i,\ y_j\neq y_i\}
\end{align}

\noindent Each anchor $x_i$ has at least one same-CUI positive in the batch (typically an alias in a different language); all remaining items in $\mathcal{B}$ serve as candidate negatives.
We then form training triplets $(x_i, x_{i^+}, x_{i^-})$, where $x_{i^+}\in \Ppairs(x_i)$ and $x_{i^-}\in \Npairs(x_i)$.
Following \SapBERTmulti{}~\cite{DBLP:conf/naacl/LiuSMBC21}, we apply in-batch hard mining and retain only \textit{hard triplets} - those in which the negative alias is close to the anchor relative to the positive, namely those satisfying 
\begin{equation}
\|\Encoder(x_i)-\Encoder(x_{i^+})\|_2 + \lambda \;\ge\; \|\Encoder(x_i)-\Encoder(x_{i^-})\|_2,
\label{equ:n}
\end{equation}
for a predefined margin $\lambda$. 
Let $\Ppairs_i$ and $\Npairs_i$ denote the indices of the positives and negatives that survive hard mining for $x_i$, and let
$S_{ij}=\cos(\Encoder(x_i),\Encoder(x_j))$.
We optimize the multi-similarity (MS) loss~\cite{DBLP:conf/cvpr/WangHHDS19}:
\begin{align}
\mathcal{L}_{\mathrm{ret}}
=&\frac{1}{N}\sum_{i=1}^{N}
\Bigg[
\frac{1}{\alpha}\log\!\left(1+\sum_{n\in \Npairs_i} \exp\big(\alpha(S_{in}-\epsilon)\big)\right) \nonumber\\
&+\frac{1}{\beta}\log\!\left(1+\sum_{p\in \Ppairs_i} \exp\big(-\beta(S_{ip}-\epsilon)\big)\right)
\Bigg] 
\end{align}
where the hyperparameters $\alpha$ and $\beta$ are scaling factors and $\epsilon$ is a similarity offset.
This objective pulls together aliases with the same CUI while pulling apart aliases with different CUIs, encouraging a clustering of the vectors based on semantic instead of syntactic similarity, and improving language-agnostic retrieval.


\subsection{Mention-Anchored Adaptation of LLM Rerankers}
\label{sec:llm}
Many biomedical mentions are ambiguous and require context for correct disambiguation: in Figure~\ref{fig:temp} the mention \textit{essoufflement} refers to the entity \textit{Dyspnea}, not to the massage technique \textit{Effleurage}, and only the surrounding text reveals that it is a symptom.
Given the top-$k$ candidate set returned by the retriever (\secref{sec:framework}), our goal is context-aware reranking without task-specific BEL annotations.

\paragraph{Why pretrained LLM rerankers.} Pretrained LLM rerankers~\cite{DBLP:journals/corr/abs-2506-05176,DBLP:journals/corr/abs-2509-25085,DBLP:journals/corr/abs-2402-03216} score retrieved candidates directly, without generating entity strings, thus avoiding the hallucinated or out-of-KB outputs of generation-based entity linking~\cite{DBLP:journals/csur/JiLFYSXIBMF23}. We therefore formulate reranking by treating the mention in context as the query and each candidate's alias as the document. 

\paragraph{The focus-mismatch problem.} Off-the-shelf LLM rerankers are trained for general query--document relevance ranking, not mention-level disambiguation, and exhibit a \textit{focus-mismatch problem}: the model may score against the most salient entity in the surrounding context instead of the target mention. Returning to the example in Fig.~\ref{fig:temp}, \textit{globules rouges} is medically salient in the context, but it is not the target of the linking decision.
Appendix~\ref{app:map} confirms this empirically - the original query--document formulation yields much weaker reranking performance than the mention-anchored formulation, indicating that off-the-shelf rerankers are not automatically aligned with the entity linking objective. 


\paragraph{Mention-anchored prompting.}
To address the focus-mismatch, we explicitly mark the target mention in the query, signaling to the reranker which token span in the surrounding context should drive the scoring decision. The reranker input $\Input\big(Q(\Mention,\Text),D(\Alias)\big)$ has two fields:
\begin{itemize}
    \item The \textbf{query} $Q(\Mention, \Text)$ is the biomedical text $\Text$ with the target mention $\Mention$ wrapped in delimiters \texttt{<tgt>...</tgt>}.
    \item The \textbf{document} $D(\Alias)$ is an alias of the candidate entity $\Entity$.
\end{itemize}
Both fields, together with a task instruction defined by the prompt template, form the reranker input, as shown in Figure~\ref{fig:temp}.

\section{Experiments}

\paragraph{Datasets.}
We evaluate \ModelName{} on four cross-lingual benchmarks---XL-BEL~\cite{DBLP:conf/acl/0001VKC20}, EMEA~\cite{10.1093/jamia/ocv037}, Patent~\cite{10.1093/jamia/ocv037}, and WikiMed-DE~\cite{DBLP:conf/wikidata/WangDS23}---covering multiple languages, domains, and KB settings. We additionally evaluate on MedMentions~\cite{DBLP:conf/akbc/MohanL19} as a large-scale English BEL transfer check, to examine whether Wikidata-derived alias supervision and mention-anchored reranking also benefit English BEL without supervised training. XL-BEL, EMEA, and Patent are evaluation-only benchmarks. For WikiMed-DE and MedMentions, we report results on their test split. Full dataset details are provided in Appendix~\ref{app:data}. 

\paragraph{Baselines.}
Following prior work on cross-lingual BEL, we compare against the closest state-of-the-art and recent systems with cross-lingual capabilities and comparable supervision assumptions, including SapBERT$_{\mbox{de}}$~\cite{mustafa-etal-2024-leveraging} and BERGAMOT~\cite{sakhovskiy-etal-2024-biomedical}. More recent state-of-the-art BEL systems, such as ANGEL~\cite{DBLP:conf/acl/KimKPLSK25}, use target-dataset supervision and are therefore evaluated separately on MedMentions. Since MedMentions is not a cross-lingual benchmark and most strong baselines are trained directly on its annotations, we report and discuss these results separately in Appendix~\ref{app:medmentions}. Full baseline details are provided in Appendix~\ref{app:baselines}. We additionally evaluate English translation baselines, where non-English mentions are first translated into English and then processed by English BEL systems. Full results and analysis are provided in Appendix~\ref{app:translation}. 

\paragraph{Experimental Setting.}
At inference time, the \ModelName{} Retriever returns the top-$k$ candidate aliases, which are then reranked by a pretrained reranker using our adaptation. Following prior work, we report Recall@1 on every baseline and benchmark. We define an \textit{unseen mention} as an evaluation mention whose exact surface string is absent from the aliases used to train our retriever. Since we discard all training aliases that exactly match evaluation mention strings, all evaluated mentions are unseen under this definition. Therefore, our results measure \textit{zero-shot} generalization to unseen mention surface forms, although the corresponding entities may still be represented during training through other aliases. Full training, inference, and evaluation details are provided in Appendix~\ref{experiment_setup}.

\paragraph{Model Choice.}
Since our work targets clinical applications involving sensitive patient data, we choose open-weight models for privacy, reproducibility, and deployment stability. We therefore instantiate our mention-anchored reranking approach with \textsc{Qwen}3-Reranker-8B in the main experiments, rather than proprietary API-based models such as GPT. Our adaptation is not specific to one LLM. Additional experiments with other pretrained LLMs are provided in Appendix~\ref{app:llm-reranker}.

\subsection{Main Results}
\label{sec:mr}
\begin{table*}[t]
\centering
\small
\setlength{\tabcolsep}{3.2pt}

\begin{tabular*}{\textwidth}{@{\extracolsep{\fill}}
l
*{10}{S[table-format=2.1]}
S[table-format=2.1]
}
\toprule
\textbf{Dataset}$\rightarrow$
& \multicolumn{10}{c}{\textbf{XL-BEL}} 
& \textbf{Avg} \\
\cmidrule(lr){2-11}
\textbf{Method}$\downarrow$
& \texttt{EN} & \texttt{ES} & \texttt{DE} & \texttt{FI} & \texttt{RU} & \texttt{TR} & \texttt{KO} & \texttt{ZH} & \texttt{JA} & \texttt{TH}
& \\
\midrule
mBERT~\cite{devlin-etal-2019-bert}
& 29.1 & 11.9 & 5.2 & 1.6 & 5.1 & 8.0 & 0.6 & 3.6 & 9.1 & 0.9
& 7.5 \\
mBART~\cite{DBLP:journals/tacl/LiuGGLEGLZ20}
& 34.8 & 23.5 & 15.0 & 11.7 & 20.4 & 24.9 & 9.0 & 8.3 & 13.1 & 11.2
& 17.2 \\
\SapBERTmulti{}~\cite{DBLP:conf/acl/0001VKC20}
& 86.2 & 48.4 & 28.5 & 19.9 & 38.4 & 35.5 & 14.8 & 14.5 & 23.1 & 14.0
& 32.3 \\
CODER~\cite{DBLP:conf/naacl/Yuan0022}
& 85.9 & 51.4 & 31.0 & 20.3 & 37.3 & 35.4 & 1.4 & 14.4 & 22.6 & 2.4
& 30.2 \\
mGENRE~\cite{DBLP:journals/tacl/CaoWPAGPZCRP22}
& 85.2 & 48.9 & 30.9 & 23.8 & 39.9 & 36.9 & 15.7 & 16.7 & 23.9 & 16.0
& 33.8 \\
Con2GEN~\cite{DBLP:conf/emnlp/ZhuQCMYX23}
& 86.6 & 52.7 & 30.8 & 24.4 & 39.8 & 40.5 & 17.4 & 17.1 & 25.6 & 19.2
& 35.4 \\
BERGAMOT~\cite{sakhovskiy-etal-2024-biomedical}
& 78.6 & 58.2 & 33.2 & 22.9 & 37.3 & 41.9 & 18.5 & 18.9 & 25.4 & 21.5
& 35.6 \\
SapBERT$_{\mbox{de}}$~\cite{mustafa-etal-2024-leveraging}
& 83.0$^{\divideontimes}$ & 57.7$^{\divideontimes}$ & 40.1$^{\divideontimes}$ & 17.0$^{\divideontimes}$ & 33.5$^{\divideontimes}$ & 42.1$^{\divideontimes}$ & 17.4$^{\divideontimes}$ & 18.8$^{\divideontimes}$ & 23.8$^{\divideontimes}$ & 20.3$^{\divideontimes}$
& 35.4$^{\divideontimes}$ \\
\midrule
\ModelName{} retriever (\textbf{ours})
& \underline{89.9} & \underline{63.7} & \underline{45.4} & \underline{31.0} & \underline{49.5} & \underline{54.5} & \underline{35.4} & \underline{35.0} & \underline{32.6} & \underline{41.2}
& \underline{47.8} \\
\ModelName{} retriever+reranker (\textbf{ours})
& \textbf{91.2} & \textbf{69.6} & \textbf{53.0} & \textbf{35.9} & \textbf{56.6} & \textbf{60.3} & \textbf{39.6} & \textbf{44.5} & \textbf{39.1} & \textbf{48.6}
& \textbf{53.8} \\
\bottomrule
\end{tabular*}

 \caption{R@1 (\%) on XL-BEL. English (\texttt{EN}), Spanish (\texttt{ES}), German (\texttt{DE}), Finnish (\texttt{FI}), Russian (\texttt{RU}), Turkish (\texttt{TR}), Korean (\texttt{KO}), Chinese (\texttt{ZH}), Japanese (\texttt{JA}), Thai (\texttt{TH}). $^{\divideontimes}$ indicates results reproduced by us. Other numbers are as reported by ~\protect\citeauthor{DBLP:conf/emnlp/ZhuQCMYX23} and ~\protect\citeauthor{sakhovskiy-etal-2024-biomedical} The best results are bold, second best are underlined. For \ModelName{}, we report the mean over three random seeds; standard deviations are provided in Appendix~\ref{app:rs}.}
\label{tab:xlbel}

\end{table*}

\begin{table*}[t]
\centering
\small
\setlength{\tabcolsep}{3.2pt}

\begin{tabular*}{\textwidth}{@{\extracolsep{\fill}}
l *{4}{S[table-format=2.1]} S[table-format=2.1] *{2}{S[table-format=2.1]} S[table-format=2.1] *{2}{S[table-format=2.1]}
}
\toprule
\textbf{Dataset}$\rightarrow$
& \multicolumn{4}{c}{\textbf{EMEA}}
& \textbf{Avg}
& \multicolumn{2}{c}{\textbf{Patent}}
& \textbf{Avg}
& \multicolumn{2}{c}{\textbf{WikiMed-DE}} \\
\cmidrule(lr){2-5}\cmidrule(lr){7-8}\cmidrule(lr){10-11}
\textbf{Method}$\downarrow$
& \texttt{ES} & \texttt{FR} & \texttt{NL} & \texttt{DE}
&
& \texttt{FR} & \texttt{DE}
&
& \texttt{DE} & \texttt{DE-wiki}\\
\midrule
mBERT~\cite{devlin-etal-2019-bert}
& 4.4 & 10.0 & 4.8 & 4.0
& 5.8
& 7.0 & 4.6
& 5.8
& 21.9$^{\divideontimes}$
& 70.1$^{\divideontimes}$ \\
mBART~\cite{DBLP:journals/tacl/LiuGGLEGLZ20}
& 35.6 & 30.9 & 29.0 & 29.9
& 31.4
& 27.2 & 29.0
& 28.1
& 19.2$^{\divideontimes}$
& 70.4$^{\divideontimes}$ \\
\SapBERTmulti{}~\cite{DBLP:conf/acl/0001VKC20}
& 58.3 & 53.4 & 55.5 & 53.6
& 55.2
& 63.5 & 61.2
& 62.4
& 59.2$^{\divideontimes}$
& 75.6 \\
CODER~\cite{DBLP:conf/naacl/Yuan0022}
& 56.9 & 53.6 & 56.9 & 53.4
& 55.2
& 64.9 & 65.8
& 65.4
& 59.2$^{\divideontimes}$
& 74.5$^{\divideontimes}$ \\
mGENRE~\cite{DBLP:journals/tacl/CaoWPAGPZCRP22}
& 56.5 & 54.8 & 49.3 & 52.7
& 53.3
& 61.1 & 58.6
& 59.9
& 26.8$^{\divideontimes}$
& 73.6$^{\divideontimes}$ \\
Con2GEN~\cite{DBLP:conf/emnlp/ZhuQCMYX23}
& 63.9 & 57.1 & 51.4 & 55.5
& 57.0
& 64.0 & 66.1
& 65.1
& \multicolumn{1}{c}{NA}
& \multicolumn{1}{c}{NA} \\
BERGAMOT~\cite{sakhovskiy-etal-2024-biomedical}
& \underline{65.7}\repro & \underline{61.0}\repro & 59.4\repro & 58.8\repro
& \underline{61.2}\repro
& 71.9\repro & 62.6\repro
& 67.3\repro
& 53.4$^{\divideontimes}$
& 77.2$^{\divideontimes}$ \\
SapBERT$_{\mbox{de}}$~\cite{mustafa-etal-2024-leveraging}
& 63.9$^{\divideontimes}$ & 58.9$^{\divideontimes}$ & 58.8$^{\divideontimes}$ & 59.1\repro
& 60.2$^{\divideontimes}$
& 71.9$^{\divideontimes}$ & 63.2$^{\divideontimes}$
& 67.6$^{\divideontimes}$
& 57.2$^{\divideontimes}$
& \textbf{80.0}\\
\midrule
\ModelName{} retriever (\textbf{ours})
& 63.3 & 60.6 & \underline{60.0} & \underline{60.2}
& 61.1
& \underline{73.0} & \underline{67.2}
& \underline{70.1}
& \underline{63.9} & 78.1\\
\ModelName{} retriever+reranker (\textbf{ours})
& \textbf{68.9} & \textbf{65.4} & \textbf{65.6} & \textbf{64.6}
& \textbf{66.1}
& \textbf{78.8} & \textbf{74.2}
& \textbf{76.5}
& \textbf{67.0} & \underline{79.8}\\ 
\bottomrule
\end{tabular*}

\caption{R@1 (\%) on EMEA, Patent and WikiMed-DE. We include two KBs for evaluation on German: (1) UMLS 2020AA (\texttt{DE}), and (2) UMLS-Wikidata~\protect\cite{mustafa-etal-2024-leveraging} (\texttt{DE-wiki}). $\divideontimes$ indicates reproduced results, while NA means the code is not fully available. Other numbers are as reported by~\protect\citeauthor{DBLP:conf/emnlp/ZhuQCMYX23} The best results are bold, second best are underlined. We report the mean over three random seeds; standard deviations are provided in Appendix~\ref{app:rs}.}
\label{tab:emaa_patent_wiki}
\end{table*}

Tables~\ref{tab:xlbel} and~\ref{tab:emaa_patent_wiki} report R@1 on XL-BEL, EMEA, Patent, and WikiMed-DE. \ModelName{} achieves the best overall performance across all four benchmarks, establishing a new state-of-the-art for cross-lingual EL.

\paragraph{Does cross-lingual alias supervision improve retrieval?}
Across the four benchmarks, the \ModelName{} retriever substantially outperforms prior methods, confirming our first hypothesis that broader cross-lingual alias coverage helps retrieval for mentions in languages underrepresented in \SapBERTmulti{}'s training data. 
The largest gains appear on XL-BEL: 47.8 average R@1 vs. 32.3 (\SapBERTmulti{}) and 35.6 (BERGAMOT). 
Improvements are consistent across all XL-BEL languages and especially strong in low-resource settings (e.g., \texttt{TR}: 54.5 vs.\ 42.1, \texttt{TH}: 41.2 vs.\ 21.5 against the previous best results). 
The improvements also transfer beyond XL-BEL to other biomedical text genres: on WikiMed-DE with UMLS 2020AA, the retriever reaches 63.9 R@1, the best result in this setting. On EMEA and Patent, where surface-form ambiguity is higher, retrieval-only gains are smaller but still competitive. Appendix~\ref{app:retriever_gain_dataset} analyzes benchmark ambiguity and explains why alias-trained retrieval helps most on XL-BEL and WikiMed-DE while EMEA and Patent require context-aware reranking. 

\paragraph{Do the gains come from cross-lingual alias coverage?}

Yes: among the 1,416 mentions improved over \SapBERTmulti{}, 82.3\% have gold CUIs with aliases in other languages, while none have only same-language Wikidata coverage (Appendix~\ref{app:error_analysis}). Broader multilingual alias coverage yields larger gains, but even concepts without any Wikidata-derived aliases in the training set still improve, suggesting that continued contrastive training also reshapes the global geometry of the embedding space, reducing the model's reliance on superficial lexical similarity. The retrieval gains are therefore driven by cross-lingual alias coverage rather than same-language alias memorization or sheer exposure to more alias strings, further supporting our first hypothesis. 

\paragraph{How much does our mention-anchored LLM reranking help beyond retrieval?}
Substantially: our pretrained LLM reranker with mention-anchored prompting adds 3-6 R@1 points across benchmarks, supporting our second hypothesis that pretrained LLM rerankers, once adapted to focus on the target mention, can provide context-aware reranking without supervised BEL annotations.  On XL-BEL it increases the average R@1 from 47.8 to 53.8, improving performance in every language. EMEA improves from 61.1 to 66.1 and Patent from 70.1 to 76.5, exceeding all prior results. On WikiMed-DE (UMLS 2020AA), reranking adds 3.1 R@1 points over the retriever. Once the correct entity is included in the candidate set, context-aware candidate comparison improves linking - especially where alias similarity alone is insufficient. 

\paragraph{When does reranking help less?}
Reranking helps less when there is little ambiguity to begin with.
Table~\ref{tab:emaa_patent_wiki} shows a special case: WikiMed-DE with the UMLS-Wikidata KB~\cite{mustafa-etal-2024-leveraging}. This KB is specifically constructed from Wikidata for WikiMed-DE, so many benchmark mentions closely match the corresponding entity aliases. Surface-form retrieval is therefore already highly effective, and the gold entity is often ranked at the top by the retriever alone. Adding the reranker yields only a marginal gain (78.1 $\rightarrow$ 79.8, +1.7).
Most real-world BEL settings rarely come with such a benchmark-specific KB. They usually link mentions to comprehensive KBs like UMLS, which provide broader entity coverage and are closer to clinical practice. These settings leave more unresolved ambiguity after retrieval, which explains the larger reranking gains on XL-BEL, EMEA, Patent, and WikiMed-DE with UMLS 2020AA.

\subsection{Ablation Study}
To understand how retrieval quality affects reranking, we vary the retrievers (BM25, mBART, BERGAMOT, and the \ModelName{} retriever) while keeping the reranker fixed. The results show that retrieval recall, R@$k$, caps the reranked R@1, so stronger retrieval leads to better overall performance. The \ModelName{} retriever performs best among the evaluated retrievers, as reported in Appendix~\ref{app:retre_affect}.

We also examine whether adding entity metadata - such as entity type or description - improves reranking. As shown in Appendix~\ref{app:entity_info}, this additional metadata does not improve performance.

Finally, three ablations in Appendix~\ref{app:llm-reranker} provide additional evidence for our second hypothesis (H2).
Appendix~\ref{app:lora} shows that mention-anchored prompting consistently outperforms lightweight supervised fine-tuning using LoRA adapters~\cite{hu_lora}. Appendix~\ref{app:map} and \ref{app:general-purpose-transfer} compare the original query--document formulation with mention-anchored prompting across LLM reranker families and model sizes, showing that explicitly marking the target mention consistently mitigates the focus-mismatch problem across different LLM rerankers. Appendix~\ref{app:maptag} analyzes mention-marker design, showing that the \texttt{<tgt>} marker performs best among those tested.

\section{Conclusion}
\ModelName{} improves cross-lingual BEL without task-specific annotations by addressing two key challenges: limited multilingual evidence for non-English retrieval, and context-aware disambiguation without supervised BEL training. The gains stem from two complementary contributions. First, we show that Wikidata-derived cross-lingual alias supervision creates shared multilingual concept neighborhoods in the embedding space, enabling zero-shot generalization to unseen mentions through cross-lingual alias bridges; this mechanism alone yields +15.5 average R@1 points on XL-BEL over \SapBERTmulti{}. Second, \ModelName{} is, to our knowledge, among the first frameworks to adapt pretrained LLM rerankers to entity linking without task-specific BEL supervision. We identify a focus-mismatch problem when off-the-shelf LLM rerankers are applied to EL and propose mention-anchored prompting as a training-free solution that consistently improves results beyond retrieval alone across all benchmarks.
\section*{Limitations}
While \ModelName{} substantially improves cross-lingual BEL without task-specific annotations, the approach has the following limitations. 

First, the reranker is constrained by the entity-side information available at inference time. In our main setting, the candidate document mainly contains the entity alias, because many KB entities lack rich textual descriptions. When multiple CUIs share near-identical metadata, such as aliases and entity types, even a strong context-aware reranker may not have sufficient candidate-side evidence to distinguish them. Enriching candidate representations with reliable descriptions or graph-based relations may improve fine-grained disambiguation, but such metadata are sparse and unevenly available across languages and KBs.


Second, following prior work~\cite{DBLP:conf/emnlp/WuPJRZ20,DBLP:conf/naacl/0003AMM22,sakhovskiy-etal-2024-biomedical}, we assume mentions are already identified. Our framework, therefore, addresses entity linking rather than end-to-end entity recognition and linking. In practical multilingual biomedical systems, mention detection can introduce additional errors, especially for low-resource languages and morphologically complex mentions. Extending \ModelName{} with robust multilingual mention detection is an important direction for future work.

\section*{Ethical Considerations and Licenses}

BioELX is intended to support biomedical entity linking research and downstream biomedical NLP applications. As with other biomedical NLP systems, incorrect entity linking predictions may affect downstream analyses if used without validation. Real-world applications in clinical decision-making settings should therefore include domain-expert review.

BioELX uses Wikidata-derived aliases as cross-lingual supervision and \textsc{Qwen}3-Reranker-8B as the main LLM reranker. Wikidata is released under the Creative Commons CC0 License. The \textsc{Qwen}3 embedding and reranking model series, including \textsc{Qwen}3-Reranker-8B, is released under the Apache 2.0 License. These uses are consistent with the stated licenses of the corresponding artifacts. Our released artifacts will include license and usage information.

Other pretrained models, datasets, and knowledge bases used in our evaluation are cited in the paper and used in accordance with their stated terms of use.


\section*{Acknowledgments}
This project was funded by the Mertelsmann Foundation gGmbH (project number: 2025034). Computational resources were provided by the HoreKa supercomputer at the Karlsruhe Institute of Technology (KIT) under project PAI00036.


\bibliography{custom}
\newpage
\appendix
\label{appendix}
\section{Notation Summary}
\begin{table}[H]
\centering
\small
\resizebox{\columnwidth}{!}{
\begin{tabular}{ll}
\toprule
\textbf{Symbol} & \textbf{Meaning} \\
\midrule
$\Text=(t_1,\ldots,t_N)$ & Input biomedical text. \\
$t_i$ & The $i$-th token in $\Text$. \\
$\Mentions$ & Set of detected mentions. \\
$\Mention$ & mention \\
$\Mention_i$ & The $i$-th mention. \\
$n_i$ & Length of mention $\Mention_i$. \\
$\KBase$ & Target biomedical knowledge base. \\
$\Entity$ & Entity in the KB. \\
$\Entity_i$ & the $i$-th Entity in the KB. \\
$f(\Mention,\Text,\KBase)$ & Entity-linking function. \\
$\Aliases(\Entity)$ & Alias set associated with entity $\Entity$. \\
$\Alias$ & Alias.\\
\midrule
$\Encoder(\cdot)$ & Retriever encoder. \\
$\cos(\cdot)$ & Cosine similarity. \\
$s_{\mathrm{ret}}(\Mention,\Alias)$ & Retrieval score between mention $\Mention$ and alias $\Alias$. \\
$\Candidate_k(\Mention)$ & Top-$k$ candidate alias set for the mention $\Mention$. \\
$k$ & Number of retrieved candidates. \\
\midrule
$\Input(Q(\Mention,\Text),D(\Alias))$ & Reranker input. \\
$Q(\Mention,\Text)$ & Mention-anchored query. \\
$D(\Alias)$ & Candidate document. \\
$\Ranker$ & Pretrained LLM reranker. \\
$s_{\mathrm{rank}}(\Mention,\Alias\mid\Text)$ & Reranking score. \\
$\Alias_{\text{final}}$ & The final alias after reranking. \\
\midrule
$\Qid$ & Wikidata item identifier. \\
$\Language$ & Language key. \\
$\Cui$ & UMLS concept unique identifier. \\
$(\Qid,\Alias,\Language)$ & Wikidata alias triple. \\
$(\Qid,\Alias,\Language,\Cui)$ & Alias tuple mapped to UMLS. \\
\midrule
$\mathcal{B}=\{(x_i,y_i)\}_{i=1}^{N}$ & Mini-batch for contrastive training. \\
$x_i$ & Alias string. \\
$y_i$ & CUI label of $x_i$. \\
$x_{i^+}$ & positive alias for $x_i$. \\
$x_{i^-}$ & negative alias for $x_i$ . \\
$\Ppairs(x_i)$ & Positive alias set for $x_i$. \\
$\Npairs(x_i)$ & Negative alias set for $x_i$. \\
$S_{ij}$ & Cosine similarity of $x_i$ and $x_j$. \\
$S_{ip}$ & Cosine similarity of $x_i$ and positive alias $x_p$. \\
$S_{in}$ & Cosine similarity of $x_i$ and negative alias $x_n$. \\
$\lambda$ & Hard-mining margin. \\
$\alpha,\beta$ & Scaling factors in MS loss. \\
$\epsilon$ & Similarity offset in MS loss. \\
$\mathcal{L}_{\mathrm{ret}}$ & Retrieval training loss. \\
\bottomrule
\end{tabular}
}
\caption{Notation used in the paper.}
\label{tab:notation}
\end{table}

\section{Experimental Setup and Implementation}
\label{experiment_setup}
\subsection{Datasets}
\label{app:data}
\begin{table*}[t]
\centering
\small
\setlength{\tabcolsep}{4pt}
\resizebox{\linewidth}{!}{
\begin{tabular}{l *{10}{r} *{4}{r} *{2}{r}*{1}{c} c}
\toprule
& \multicolumn{10}{c}{\textbf{XL-BEL}} & \multicolumn{4}{c}{\textbf{EMEA}} & \multicolumn{2}{c}{\textbf{Patent}}&\multicolumn{1}{c}{\textbf{WikiMed-DE}}&\multicolumn{1}{c}{\textbf{MedMentions}}  \\
\cmidrule(lr){2-11}\cmidrule(lr){12-15}\cmidrule(lr){16-17}\cmidrule(lr){18-18}\cmidrule(lr){19-19}
& EN & ES & DE & FI & RU & TR & KO & ZH & JA & TH
& ES & FR & NL & DE
& FR & DE 
& DE
& EN\\
\midrule
Mentions  
& 1000 & 1000 & 1000 & 1000 & 1000 & 1000 & 1000 & 1000 & 1000 & 1000
& 432 & 431 & 434 & 425
& 342 & 348 & 41,120 
& 40,116\\
Entities  
& 807 & 889 & 908 & 816 & 816 & 694 & 829 & 875 & 822 & 704
& 295 & 304 & 304 & 299
& 223 & 223 &9,123 
& 8,457\\
\bottomrule
\end{tabular}}
\caption{Dataset statistics for XL-BEL, EMEA, Patent, WikiMed-DE and MedMentions across languages. For WikiMed-DE and MedMentions, we only report the statistics for their test set because we do not use them for training.}
\label{tab:data_stats}
\end{table*}
We evaluate \ModelName{} on five biomedical entity linking benchmarks, covering multiple languages, diverse domains, and different settings (i.e., from single sentences to paragraphs). Table~\ref{tab:data_stats} summarizes the number of mentions and unique entities per language for each benchmark. 
\begin{itemize}
    \item \textbf{XL-BEL}~\cite{DBLP:conf/acl/0001VKC20}: a multilingual benchmark covering 10 typologically diverse languages: English (EN), Spanish (ES), German (DE), Finnish (FI), Russian (RU), Turkish (TR), Korean (KO), Chinese (ZH), Japanese (JA), Thai (TH). Each language contains 1,000 annotated sentences extracted from Wikipedia, each with a single annotated mention. 
    \item \textbf{EMEA}~\cite{10.1093/jamia/ocv037}: the text of this benchmark comes from the European Medicines Agency. It covers four languages (ES, FR, NL, DE) with sentence-level contexts and UMLS-linked mentions.
    \item \textbf{Patent}~\cite{10.1093/jamia/ocv037}: text of this benchmark comes from the European Patent Office. It contains two languages, German and French, also annotated with UMLS-linked mentions.\footnote{We follow the same data source as EMEA; the ``Patent'' subset is reported separately due to its distinct domain.}
    \item \textbf{WikiMed-DE}~\cite{DBLP:conf/wikidata/WangDS23}: a large-scale, silver-standard German benchmark constructed from Wikipedia with UMLS-aligned entities; we report results on its test set containing 41,120 mentions.
    \item \textbf{MedMentions}~\cite{DBLP:conf/akbc/MohanL19}
    an English biomedical entity linking benchmark constructed from PubMed abstracts, with expert-linked mentions to UMLS CUIs. Since MedMentions is not cross-lingual and most strong baselines are trained on its annotations, we use it only as an additional English BEL transfer check. It covers a broad range of biomedical concepts and provides document-level contexts; we evaluate on the standard test split.
\end{itemize}

\subsection{Knowledge Bases}
\label{app:kb}
To ensure comparability, we follow prior work~\cite{DBLP:conf/acl/0001VKC20,DBLP:conf/emnlp/ZhuQCMYX23,mustafa-etal-2024-leveraging} and use the same knowledge bases for evaluation:
\begin{itemize}
    \item \textbf{UMLS 2020AA version}~\cite{DBLP:journals/nar/Bodenreider04}: we use the full release, containing 4,281,184 entities. We use this KB to evaluate EMEA, Patent and WikiMed-DE (UMLS setting).  
    \item \textbf{UMLS 2020AA Level-0 subset}: the subset of UMLS 2020AA restricted to source vocabularies with UMLS restriction level 0, containing 3,377,062 entities. For the ablation study conducted on EMEA and Patent, we use this knowledge base (Appendix~\ref{app:llm-reranker}).
    \item \textbf{UMLS 2017AA version}~\cite{DBLP:journals/nar/Bodenreider04}: we use the full release of this version, containing 2,595,474 entities to evaluate MedMentions, which was annotated using this UMLS version.
    \item \textbf{UMLS-399931}~\cite{DBLP:conf/acl/0001VKC20}: the KB provided with the XL-BEL benchmark. It is a subset of UMLS 2020AA, containing 62,094 entities and 399,931 aliases. We use this KB to evaluate XL-BEL.
    \item \textbf{UMLS--Wikidata}~\cite{mustafa-etal-2024-leveraging}: a German KB constructed for WikiMed-DE by collecting German Wikidata items that are linked to a UMLS CUI. We use this KB only for evaluating WikiMed-DE under the UMLS--Wikidata setting.
\end{itemize}

\subsection{Baselines}
\label{app:baselines}
Following prior studies~\cite{sakhovskiy-etal-2024-biomedical,DBLP:conf/emnlp/ZhuQCMYX23,mustafa-etal-2024-leveraging}, we compare \ModelName{} against representative baselines covering dense retrieval, alias-based biomedical retrieval, generative linking, KB-enhanced retrieval, and supervised biomedical EL. For the cross-lingual benchmarks, we compare against multilingual BEL systems. For MedMentions, we follow the evaluation setting in~\citet{DBLP:conf/acl/KimKPLSK25} and compare against strong systems trained specifically on MedMentions.
\paragraph{Baselines for multilingual benchmarks.}
\begin{itemize}
    \item \textbf{mBERT}~\cite{devlin-etal-2019-bert}: a multilingual Transformer encoder used as a dense retriever. Mentions and entity names are encoded into the same vector space, and linking is performed by nearest-neighbor search. Unlike biomedical alias-trained models, mBERT is not specialized for biomedical synonym alignment.
    \item \textbf{mBART}~\cite{DBLP:journals/tacl/LiuGGLEGLZ20}: a multilingual sequence-to-sequence model used as an encoder baseline for dense retrieval. We encode mentions and entity names with the same model and perform nearest-neighbor search in the embedding space.
    \item \textbf{\SapBERTmulti{}}~\cite{DBLP:conf/acl/0001VKC20}: a multilingual biomedical retriever trained with a contrastive objective on UMLS aliases. It learns to align aliases that share the same CUI and serves as the main alias-based retrieval baseline. \ModelName{} builds on this representation but expands the cross-lingual alias supervision with Wikidata-derived aliases.
    \item \textbf{CODER}~\cite{DBLP:conf/naacl/Yuan0022}: a biomedical encoder that incorporates knowledge signals such as concept structure into contrastive representation learning. It is used as a dense retrieval baseline in the same nearest-neighbor linking setup.
    \item \textbf{mGENRE}~\cite{DBLP:journals/tacl/CaoWPAGPZCRP22}: a multilingual generative entity linking model. Instead of retrieving candidates with dense similarity, it generates entity identifiers or names through constrained decoding.
    
    \item \textbf{Con2GEN}~\cite{DBLP:conf/emnlp/ZhuQCMYX23}: a cross-lingual biomedical entity linking framework that combines \SapBERTmulti{} retrieval with controllable generation based on mGENRE. It injects KB-side attributes (e.g., entity type) into the generation process to better disambiguate lexically ambiguous mentions. We do not report Con2GEN on WikiMed-DE because there is no fully reproducible code for this baseline.
    
    \item \textbf{BERGAMOT}~\cite{sakhovskiy-etal-2024-biomedical}: a cross-lingual biomedical entity linking approach that incorporates structured UMLS knowledge (e.g., graph) for representation learning.
    
    \item \textbf{SapBERT$_{\mbox{de}}$}~\cite{mustafa-etal-2024-leveraging}: A German-adapted model fine-tuned based on \SapBERTmulti{} for WikiMed-DE. We also report its results on other benchmarks to compare with our model, \ModelName{}.
\end{itemize}

All baseline numbers in our main paper are reported by~\citeauthor{DBLP:conf/emnlp/ZhuQCMYX23} and ~\citeauthor{sakhovskiy-etal-2024-biomedical} except the numbers annotated with the symbol $^{\divideontimes}$. For the numbers annotated with $^{\divideontimes}$, we run the released checkpoints under the same evaluation protocol.

\paragraph{Baselines for MedMentions.}
We additionally evaluate on MedMentions as an English BEL transfer check. Except for \SapBERTmulti{}, the following baselines are trained on the MedMentions training set, so the comparison is not supervision-matched with \ModelName{}.

\begin{itemize}
    \item \textbf{\SapBERTmulti{}}: the same UMLS-alias-trained biomedical retriever used in the multilingual experiments. It provides a direct, retrieval-only comparison to assess whether \ModelName{} improves over the original alias-based representation.
    \item \textbf{GenBioEL}~\cite{DBLP:conf/naacl/Yuan0022}: a supervised generative BEL model that casts linking as sequence generation, producing the entity name/identifier with constrained decoding over a KB-derived vocabulary space.
    \item \textbf{ArboEL}~\cite{DBLP:conf/naacl/0003AMM22}: a supervised retrieve--rerank framework trained on MedMentions, which trains a bi-encoder for candidate retrieval and a cross-encoder for reranking using mention context and entity text.
    \item \textbf{Prompt-BioEL}~\cite{DBLP:conf/aaai/XuCH23}: a supervised retrieve--rerank method. It trains a bi-encoder retriever on mentions and entity names, and then applies a prompt-tuned reranker that jointly encodes the mention context with candidate entities to model cross-entity interaction.
    \item \textbf{GenBioEL+ANGEL}~\cite{DBLP:conf/acl/KimKPLSK25}: a biomedical entity linking system fine-tuned based on GenBioEL with hard negative optimization.
\end{itemize}
All numbers for these approaches are reported in the original papers. These MedMentions baselines are optimized directly on the target benchmark, whereas \ModelName{} does not use MedMentions supervision. We therefore use MedMentions only as an additional English transfer check rather than as the primary setting for our cross-lingual BEL claims.

\subsection{Implementation Details}
\paragraph{Continued Contrastive Training of \ModelName{} retriever.} 
We follow the \SapBERTmulti{} training setup and construct positive alias pairs within each CUI. To avoid over-representing CUIs with many aliases and to keep training efficient, we randomly sample at most 50 positive pairs per CUI. We use the margin $\lambda$ = 0.2 for online hard example mining, and represent each input using the \texttt{[CLS]} embedding. 
Training runs for 5 epochs with a batch size of 256, a maximum sequence length of 25, and a learning rate of 2e-5, and is optimized using AdamW with a weight decay of 0.01.

\paragraph{Knowledge base clarification.} 
Following the format of UMLS-399931, we extract every (CUI, name) pair from \texttt{MRCONSO.RRF} across all source vocabularies and languages, store each pair as one \texttt{CUI||name} line, and remove duplicate pairs.  We build two versions of this KB: (1) a \emph{lowercased} version, in which all names are lowercased before de-duplication (about 666\,MB, 12,671,936 lines), and (2) a \emph{case-preserving} version that keeps the original casing of the names (about 681\,MB, 13,080,061 lines). Both versions cover the same 4,281,184 CUIs. Since some baseline results in our main table are taken from \citet{DBLP:conf/emnlp/ZhuQCMYX23}, whose reported KB size (12,671,936 names) is identical to that of our lowercased version, we report our main results (Table~\ref{tab:emaa_patent_wiki}) on the lowercased version for a fair comparison. Results on the case-preserving version are given in Table~\ref{tab:emaa_patent_wiki_cased}
  
\paragraph{Inference.} At inference time, we use \ModelName{} Retriever to generate the top-$k$ candidate aliases with associated CUIs for each mention. We set $k=64$ for XL-BEL and $k=16$ for all other benchmarks. We then apply the \textsc{Qwen}3‑Reranker‑8B model~\cite{DBLP:journals/corr/abs-2506-05176} to score each (query, candidate‑document) pair. The reranker operates in batches over candidates, producing a scalar relevance score for each candidate alias. Candidate aliases are re-sorted according to the reranker scores, and the final prediction is the CUI associated with the highest-scoring alias.

\paragraph{Metrics.}
Following prior work on EL~\cite{logeswaran-etal-2019-zero,DBLP:conf/emnlp/WuPJRZ20}, we report Recall@$k$ (R@$k$), i.e., the fraction of mentions for which the gold CUI appears among the top-$k$ predictions. When multiple candidates receive the same highest score, we treat all tied candidates as top-1 predictions. A mention is counted as correct at R@1 if its gold CUI is among them. We apply the same evaluation procedure to all baselines.

\subsection{Reproducibility and Computational Cost}
\label{app:reproducibility}

To support reproducibility, we will release the full codebase upon publication. This includes scripts for processing Wikidata dumps and mapping Wikidata aliases to UMLS CUIs via Wikidata properties, training the \ModelName{} Retriever, and running inference with both the retriever and the LLM reranker. We will also release the processed Wikidata-derived aliases and upload the trained \ModelName{} Retriever checkpoint to Hugging Face.

All main experiments are run on a machine with 8 NVIDIA A100 80GB GPUs. Training the retriever takes approximately 45 minutes per epoch. Processing Wikidata-derived aliases is CPU-based and does not require GPUs. At inference time, retrieval and reranking have different computational profiles. The retriever requires GPU computation to encode mentions and KB aliases, but KB alias embeddings can be precomputed once and reused across runs. This one-time preprocessing cost scales with the KB size: encoding the XL-BEL KB takes approximately 3 minutes, whereas encoding the full UMLS 2020 release with millions of entity names and aliases takes approximately 2 hours.


At inference time, the \ModelName{} Retriever returns the top-$k$ candidates per mention, which are reranked using \textsc{Qwen}3-Reranker-8B. We set $k=64$ for XL-BEL and $k=16$ for all other benchmarks. On a single NVIDIA A100 80GB GPU, reranking XL-BEL (10,000 mentions; 640,000 mention--candidate pairs) takes approximately 1 hour and 44 minutes. For EMEA (1,722 mentions; 27,552 pairs) and Patent (690 mentions; 11,040 pairs), reranking takes approximately 8 and 7 minutes, respectively, on the same GPU. For WikiMed-DE (41,120 mentions; 657,920 pairs), reranking takes approximately 6 hours and 40 minutes on a single NVIDIA A40 GPU. Overall, reranking cost depends on both the number of mention--candidate pairs and the input sequence length; reducing $k$ can therefore substantially lower inference time.





\section{Wikidata-Derived Alias Supervision}
\subsection{Statistics}
\label{app:data-stat}
This section provides detailed statistics on the scale, processing stages, and language-specific UMLS concept coverage of the Wikidata-derived aliases used to train the \ModelName{} retriever.

\paragraph{Alias construction statistics.}
Table~\ref{tab:wikidata_overall} reports statistics at three stages of constructing the Wikidata-derived alias set. The \textit{Raw Wikidata extraction} block shows the number of aliases, QIDs, and languages obtained directly from the Wikidata dump before any biomedical filtering. The \textit{After CUI mapping} block reports the subset of Wikidata items that can be linked to UMLS through the \texttt{P2892} property, together with the resulting number of CUI-linked aliases, QIDs, CUIs, and languages. The \textit{After removing benchmark aliases} block provides the final dataset after decontamination, with aliases that exactly match evaluation mention strings removed. We also report the average and median numbers of languages per CUI in the final training set, along with the end-to-end processing time.

\begin{table}[h]
\centering
\small
\setlength{\tabcolsep}{4pt}
\resizebox{\columnwidth}{!}{
\begin{tabular}{l r}
\toprule
\textbf{Statistic} & \textbf{Value} \\
\midrule
\multicolumn{2}{l}{\textit{Raw Wikidata extraction}} \\
Wikidata dump version & 2025-12-01 \\
Dump file used & \texttt{wb\_items\_per\_site.sql} \\
Dump file size & 1.69 GB \\
Raw aliases extracted & 96,593,429 \\
Raw unique QIDs extracted & 33,696,395 \\
Raw languages covered & 950 \\
\midrule
\multicolumn{2}{l}{\textit{After CUI mapping}} \\
QIDs with P2892 annotation & 731,652 \\
CUI-linked aliases after mapping & 3,834,319 \\
Unique QIDs after CUI mapping & 534,156 \\
Unique CUIs after mapping & 537,848 \\
Languages covered after mapping & 597 \\
\midrule
\multicolumn{2}{l}{\textit{After removing benchmark aliases}} \\
Total aliases & 3,823,814 \\
Unique QIDs & 534,131 \\
Unique CUIs & 537,823 \\
Languages covered & 597 \\
Avg. languages per CUI & 7 \\
Median languages per CUI & 7 \\
English aliases coverage & 6.34\% \\
\midrule
Processing time (end-to-end) & 13.83 Min \\
Workstation & CPU only \\
\bottomrule
\end{tabular}}
\caption{Overall statistics of the Wikidata-derived alias extraction, CUI mapping, decontamination, and final training set construction pipeline.}
\label{tab:wikidata_overall}
\end{table}

\begin{figure}[h]
\centering
\includegraphics[width=\columnwidth]{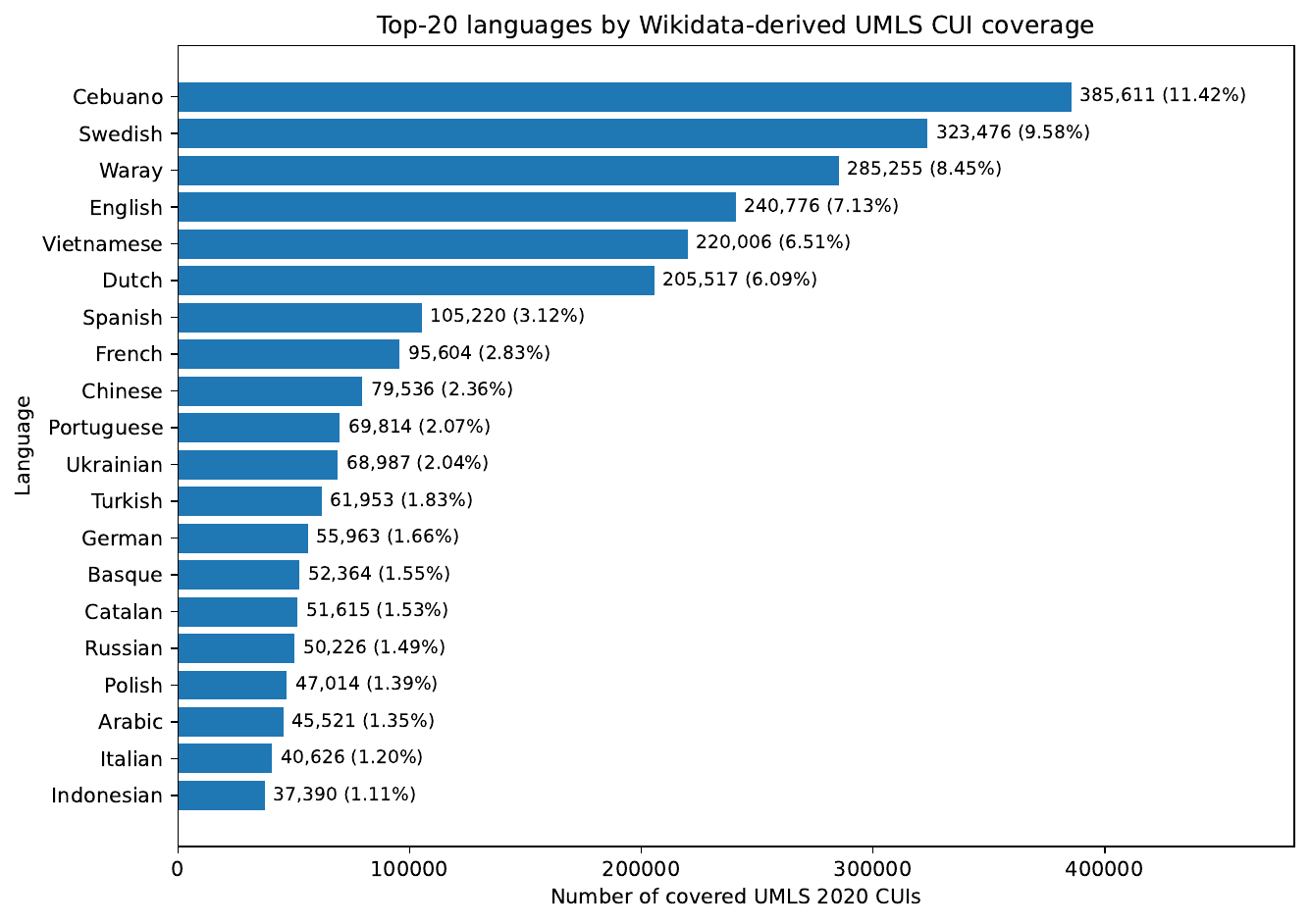}
\caption{Top 20 languages by Wikidata-derived coverage of UMLS 2020 CUIs. Each bar reports the number of UMLS CUIs associated with at least one Wikidata alias in the corresponding language. Percentages are calculated relative to all $3{,}377{,}062$ CUIs in UMLS 2020. Since a CUI may have aliases in multiple languages, language-specific coverage counts are not mutually exclusive.}
\label{fig:wikidata-cui-coverage}
\end{figure}

\paragraph{UMLS concept coverage across languages.}
UMLS has multiple releases. We measure concept coverage against UMLS 2020, which is used by our main evaluation benchmarks. Before benchmark-alias removal, $537{,}848$ of the $3{,}377{,}062$ UMLS 2020 CUIs are linked to at least one Wikidata item, corresponding to an overall CUI coverage of $15.9\%$. Figure~\ref{fig:wikidata-cui-coverage} shows the 20 languages with the highest coverage. Among the benchmark languages, English covers $240{,}776$ CUIs ($7.13\%$), Dutch $205{,}517$ ($6.09\%$), Spanish $105{,}220$ ($3.12\%$), French $95{,}604$ ($2.83\%$), Chinese $79{,}536$ ($2.36\%$), Turkish $61{,}953$ ($1.83\%$), German $55{,}963$ ($1.66\%$), and Russian $50{,}226$ ($1.49\%$). These language-specific sets overlap because the same CUI may have aliases in multiple languages.

Coverage varies substantially across languages and may affect generalization to broader BEL settings. However, coverage alone does not determine performance: alias quality, lexical variation, and benchmark difficulty also influence retrieval accuracy. Note that the percentages reported here measure CUI coverage relative to all UMLS 2020 concepts, whereas the alias-language percentages in Table~\ref{tab:wikidata_overall} measure each language's share of the extracted alias collection.

\subsection{Adaptation to Other Biomedical Resources}
\label{app:adapt_wiki_ontology}
We group aliases by Wikidata QID, which provides a common key for linking aliases and external biomedical identifiers within Wikidata. In our current setup, biomedical filtering is performed using the UMLS CUI property (P2892). The same procedure can be adapted to other resources by using their corresponding Wikidata properties, e.g., the MeSH descriptor ID (P486).

An external ontology is not strictly required for biomedical filtering. When mappings to resources such as UMLS or MeSH are unavailable, users can instead identify relevant Wikidata entities through internal \textit{instance of} (P31) and \textit{subclass of} (P279) relations, using domain-specific classes such as diseases, symptoms, genes, or proteins. As a feasibility check, we queried entities associated with the gene class (Q7187) and retrieved 1,296,824 gene-related Wikidata items whose multilingual aliases can be grouped by QID. This experiment is not used in our main training pipeline, but it illustrates that domain-specific alias supervision can also be constructed directly from Wikidata's class hierarchy without relying on an external ontology.

As auxiliary evidence for transfer beyond UMLS-based evaluation, we evaluate \ModelName{} on BC5CDR~\citep{DBLP:journals/biodb/LiSJSWLDMWL16}, whose target ontology is MeSH. Without BC5CDR-specific training data or MeSH-derived alias supervision, the BioELX retriever achieves 89.4 R@1, compared with 88.6 for SapBERT under the same setting. This result provides direct evidence that the learned representation transfers to a MeSH-based entity linking setting. 

Scripts for UMLS-, MeSH-, and Wikidata-class-based filtering are provided in our code repository.

\section{Diagnostic Experiments}
\label{app:context_retrieval}

As an initial diagnostic experiment, we test whether context can be directly incorporated into an alias-trained retriever. We use the off-the-shelf \SapBERTmulti{} checkpoint\footnote{\url{https://huggingface.co/cambridgeltl/SapBERT-UMLS-2020AB-all-lang-from-XLMR}.} and compare two retrieval inputs: the mention string alone, and the mention concatenated with its surrounding context. In both settings, the retrieval index is constructed from the same entity names and aliases, and only the input changes. Since we use a fixed pretrained checkpoint and do not train any model in this experiment, the retrieval results are deterministic and do not involve random seeds.

As shown in Table~\ref{tab:context_retrieval}, naively appending context severely degrades retrieval at all cutoffs. On EMEA, R@1 drops from 55.2 to 5.3 and R@64 from 80.1 to 22.5. On Patent, R@1 drops from 62.4 to 3.2 and R@64 from 90.3 to 24.8. These results confirm that alias-trained retrievers are sensitive to non-alias context tokens. This supports our design choice of keeping retrieval at mention level and introducing context at the reranking stage.

\begin{table}[t]
\centering
\small
\resizebox{\columnwidth}{!}{
\begin{tabular}{llrrrrr}
\toprule
Dataset & Input & R@1 & R@8 & R@16 & R@32 & R@64 \\
\midrule
\multirow{2}{*}{EMEA}
& Mention only      & 55.2 & 71.2 & 76.3 & 79.2 & 80.1 \\
& Mention + context & 5.3  & 14.9 & 18.8 & 22.5 & 22.5 \\
\midrule
\multirow{2}{*}{Patent}
& Mention only      & 62.4 & 79.8 & 82.9 & 86.8 & 90.3 \\
& Mention + context & 3.2  & 10.8 & 14.4 & 18.3 & 24.8 \\
\bottomrule
\end{tabular}
}
\caption{Effect of naively appending context to \SapBERTmulti{} retrieval on EMEA and Patent. Results are R@$k$ (\%).}
\label{tab:context_retrieval}
\end{table}

\section{LLM Reranker Details and Analysis}
\label{app:llm-reranker}

\subsection{Reranker Motivation and Selection}
\label{app:ranker_selection}
\paragraph{Why LLM rerankers?} For context-aware reranking without task-specific BEL annotations, we adopt LLM rerankers rather than generation-based LLMs. Generation-based entity linking requires the model to produce an entity name or identifier, which can lead to hallucinated or out-of-KB outputs~\cite{DBLP:journals/csur/JiLFYSXIBMF23}, especially given the diversity of biomedical aliases. In contrast, LLM rerankers score query--document pairs directly, using a relevance score, confidence score, or similarity score depending on the model.  This allows us to rank candidate entities without generating entity strings.

\paragraph{LLM Reranker Selection Criteria}
We select rerankers according to three practical criteria. First, given our target clinical use cases, we prioritize models that can be deployed locally. Patient data are confidential, and clinical data-use agreements often prohibit transmitting them to external APIs. Externally hosted models, such as GPT and Claude, may also change after updates or become unavailable, creating risks for reproducibility and long-term deployment~\cite{Hager2024}. We therefore focus on open-weight models with fixed checkpoints, enabling local deployment, reproducibility, and long-term stability.

Second, the reranker should provide a relevance score, confidence score, or similarity score for mentions and their retrieved candidates, rather than generating entity strings. This scoring-based formulation avoids generation-based failure modes such as hallucinated or out-of-KB outputs. 

Third, it should support multilingual ranking, which is essential for cross-lingual BEL. 

Representative rerankers satisfying these criteria include \textsc{Qwen}3-Ranker~\cite{DBLP:journals/corr/abs-2506-05176}, Jina-Reranker~\cite{DBLP:journals/corr/abs-2509-25085}, and BGE-Reranker~\cite{DBLP:journals/corr/abs-2402-03216}.

\paragraph{Why Qwen3-Reranker-8B in main experiments.} We use \textsc{Qwen}3-Reranker-8B as the main reranker in our primary experiments because it provides a strong open-source multilingual ranking interface and performs best among the evaluated rerankers in our transfer analysis. Appendix~\ref{app:map} shows that mention anchoring improves all evaluated rerankers, while \textsc{Qwen}3-Reranker-8B obtains the strongest overall reranking results. We therefore use it for the main tables and use Jina-Reranker and BGE-Reranker as additional checks of whether mention anchoring transfers across reranker architectures.

\subsection{Scoring Function Variants for Evaluated Rerankers}
\label{app:scoring-variants}

Mention-anchored prompting is independent of how the LLM reranker produces its scalar relevance score. Different open-source rerankers expose different scoring interfaces, summarized below. In all variants, the prompt format is identical: the query $Q(\Mention, \Text)$ contains the biomedical text with the target mention wrapped in \texttt{<tgt>...</tgt>}, and the document $D(\Alias)$ contains the candidate entity alias.

\paragraph{Confidence-based scoring function.}
\textsc{Qwen}3-Reranker~\cite{DBLP:journals/corr/abs-2506-05176} is trained to output a binary ``yes'' or ``no'' decision indicating whether the document is relevant to the query. Following~\citet{DBLP:journals/corr/abs-2506-05176}, the relevance score is the model's confidence in the ``yes'' token, obtained by extracting the logits over the vocabulary at the answer position, restricting to the ``yes'' and ``no'' tokens, applying softmax, and taking the probability of ``yes'':
\begin{equation}
\resizebox{.91\linewidth}{!}{$
            \displaystyle
s_{\mathrm{rank}}^{\mathrm{conf}}(\Mention, \Alias \mid \Text)
=\text{Softmax}\Big(l(\text{yes}\mid Q,D),l(\text{no}\mid Q,D)\Big)_{\text{yes}}
$}.
\end{equation}
where $l(\text{yes}\mid Q,D)$ and $l(\text{no}\mid Q,D)$ are logits for the next token being ``yes'' or ``no'' conditioned on the concatenated query--document input, respectively. This is the scoring form used in our main experiments.

\paragraph{Similarity-based scoring function.}
Jina-Reranker-v3~\cite{DBLP:journals/corr/abs-2509-25085} places the query and all candidate documents in a single context window and extracts contextual embeddings from designated special-token positions. A two-layer MLP projects these embeddings into a shared ranking space, and the relevance score is the cosine similarity between the projected query embedding $\mathbf{q}$ for $Q$ and the projected document embedding $\mathbf{d}$ for $D$:
\begin{equation}
s_{\mathrm{rank}}^{\mathrm{sim}}(\Mention, \Alias \mid \Text) 
= 
\cos\!\big(\mathbf{q}, \mathbf{d}\big) 
\in [-1, 1].
\end{equation}

\paragraph{Logit-based scoring function.}
BGE-Reranker-v2-Gemma~\cite{DBLP:journals/corr/abs-2402-03216} is a cross-encoder that jointly encodes the query--document pair and produces a single scalar relevance logit:
\begin{equation}
s_{\mathrm{rank}}^{\mathrm{logit}}(\Mention, \Alias \mid \Text)
=
\ell_{\text{yes}}\!\big(Q(\Mention, \Text), D(\Alias)\big)
\in \mathbb{R},
\end{equation}
where $\ell_{\text{yes}}$ is the model's logit for the ``yes'' token. Higher logits indicate higher relevance. We use the raw logits without sigmoid normalization, since within-ranker ordering is sufficient for sorting candidates.

\subsection{Mention-Anchored Prompting versus Lightweight Supervised Fine-Tuning}
\label{app:lora}
As discussed in Section~\ref{sec:llm}, off-the-shelf LLM rerankers can be distracted by salient context tokens rather than the target mention span, leading to incorrect linking decisions and limiting their direct applicability to entity linking. Recent work also highlights the value of training-free mechanisms for focusing LLMs on task-relevant evidence~\cite{zhong2026focus}. For EL, the corresponding challenge is mention-level focus: rerankers must be adapted to score candidates with respect to the target mention rather than the context as a whole. We examine whether this adaptation requires supervised parameter updates or whether it can be achieved through a training-free input refinement. We conduct this ablation with \textsc{Qwen}3-Reranker, the main reranker used in our main paper, and compare two adaptation strategies: (i) \emph{refining} the input with an explicit mention-anchored prompt, as shown in Figure~\ref{fig:temp}, and (ii) \emph{fine-tuning} the reranker on a small Wikipedia--Wikidata dataset to better follow the entity-linking instruction. Table~\ref{tab:ablation_Qwen_prompt} compares three variants:

\begin{enumerate}
    \item Vanilla \textsc{Qwen} with its default template (\textit{Orig.\ prompt})
    \item Vanilla \textsc{Qwen} with our mention-anchored prompt (\textit{Refined prompt}), and 
    \item \textsc{Qwen} further fine-tuned using our supervision while keeping the same mention-anchored prompt (\textit{Refined prompt + FT}). 
\end{enumerate}

\begin{table}[t]
\centering
\small
\setlength{\tabcolsep}{5pt}
\begin{tabular}{lccc}
\toprule

& XL-BEL 
& EMEA
& Patent \\
\midrule
\textit{Orig.\ prompt} & 45.6 & 35.8 & 54.4\\
\textit{Refined prompt} 
& \textbf{54.8} & \textbf{63.2} & \textbf{74.0} \\
\textit{Refined prompt + FT}
& 54.0 & 61.5 & 71.6 \\
\bottomrule
\end{tabular}
\caption{Ablation of prompt design vs.\ fine-tuning for the \textsc{Qwen}3-Reranker. All variants rerank the same top-$k$ candidate sets produced by the same retriever (fixed $k$), and differ only in the input prompt/template and whether supervised fine-tuning is applied. Metrics are R@1 (\%).}
\label{tab:ablation_Qwen_prompt}
\end{table}

For fine-tuning, we construct a lightweight dataset from Wikipedia and Wikidata with $1000$ instances. Each instance contains one positive entity alias and $32$ negative entity aliases. We fine-tune the reranker with parameter-efficient adaptation using LoRA~\cite{hu_lora} with rank $r=16$, a learning rate of $1\times10^{-4}$, and one training epoch. We vary the language used for fine-tuning by creating three $1000$-instance subsets: German-only, Thai-only, and a mixture across all XL-BEL languages. Among the fine-tuned variants, German performs best, but it still underperforms the non-fine-tuned baseline. To test whether the large number of negatives harms learning, we reduce the number of negatives to $7$, but the results remain largely unchanged. We also try alternative learning-rate schedules, including smaller learning rates, without improving over the baseline.

As shown in Table~\ref{tab:ablation_Qwen_prompt}, the mention-anchored prompt yields the strongest performance, while additional fine-tuning degrades the results of vanilla \textsc{Qwen} with our mention-anchored prompt. We attribute this to two factors. First, due to limited compute, we use LoRA instead of full-model fine-tuning: its restricted update capacity may be insufficient to induce the behavioral changes required for entity linking. Second, the fine-tuning instances are constructed from retrieval outputs. This dataset is small and may inherit retriever-specific biases, creating a distribution mismatch with evaluation data and limiting generalization. We expect that fine-tuning could become beneficial with a larger and more diverse supervision set and with reduced retrieval-induced bias; improving data selection and coverage is an important direction for future work.

\subsection{Does Mention-Anchored Prompting Transfer Across Rerankers and Ranking Interfaces?}
\label{app:map}
Having shown that mention-anchored prompting is more effective than lightweight fine-tuning for \textsc{Qwen}3-Reranker, we next examine whether this adaptation transfers beyond a single reranker and ranking interface.

\paragraph{Transfer across rerankers.}
Table~\ref{tab:ranker_anchor_transfer} evaluates whether mention-anchored prompting improves reranking across different reranker families and model sizes. We conduct this analysis on EMEA and Patent because they are challenging benchmarks for alias-based retrieval. The same mention string can correspond to different gold entities depending on context, making them suitable for testing context-aware disambiguation. For example, in EMEA, the mention \textit{Chemotherapie} is annotated with two different gold CUIs, C0392920 (Chemotherapy Regimen) and C0013216 (Pharmacotherapy), depending on the surrounding text.

We compare the original query--document formulation with our mention-anchored input for Jina-Reranker-V3\footnote{jina-reranker-v3: \url{https://huggingface.co/jinaai/jina-reranker-v3}}, BGE-Reranker\footnote{bge-reranker-v2-gemma: \url{https://huggingface.co/BAAI/bge-reranker-v2-gemma}}, and \textsc{Qwen}3-Reranker\footnote{Qwen3-Reranker-8B: \url{https://huggingface.co/Qwen/Qwen3-Reranker-8B}}. Our mention anchoring improves all rerankers on both benchmarks. The average gains are +7.09 R@1 on EMEA and +8.70 on Patent for Jina, +3.60 and +4.21 for BGE, +3.12 and +8.40 for \textsc{Qwen}3-Reranker-0.6B, +12.50 and +19.89 for \textsc{Qwen}3-Reranker-4B, and +18.58 and +22.02 for \textsc{Qwen}3-Reranker-8B, respectively.

These results suggest that mention-anchored prompting is broadly useful for adapting rerankers trained for general query--document relevance ranking to context-aware entity linking. At the same time, model capacity and the ranking interface both affect the magnitude of the gain. The improvement is largest for \textsc{Qwen}3-Reranker-8B, which combines larger model capacity with an instruction interface that allows us to explicitly steer the reranker to score candidates with respect to the marked target mention rather than generic query--document relevance. We therefore use \textsc{Qwen}3-Reranker-8B as the main reranker in our primary results, while using the other rerankers as additional checks of whether mention anchoring transfers across ranking architectures.

Since these rerankers are used as frozen pretrained models and no task-specific training is performed, this ablation does not involve training random seeds.

\begin{table*}[h]
\centering
\small
\setlength{\tabcolsep}{3.5pt}
\resizebox{\textwidth}{!}{
\begin{tabular}{lllcccccc cccc}
\toprule
\textbf{Dataset}$\rightarrow$ & &
& \multicolumn{6}{c}{\textbf{EMEA}} 
& \multicolumn{4}{c}{\textbf{Patent}} \\
\cmidrule(lr){4-9}
\cmidrule(lr){10-13}
\textbf{Reranker}$\downarrow$ & Model size & \textbf{Input}$\downarrow$ 
& \texttt{ES} & \texttt{FR} & \texttt{NL} & \texttt{DE} & \textbf{Avg.} & \textbf{Gain}
& \texttt{FR} & \texttt{DE} & \textbf{Avg.} & \textbf{Gain} \\
\midrule
\multirow{2}{*}{\href{https://huggingface.co/jinaai/jina-reranker-v3}{jina-reranker-v3}}
& \multirow{2}{*}{0.6B}
& Original 
& 47.92 & 47.33 & 45.62 & 44.94 & 46.45 & -- 
& 51.46 & 47.41 & 49.44 & -- \\
&
& Ours
& 55.56 & 54.76 & 53.23 & 50.59 & 53.54 & +7.09
& 60.53 & 55.75 & 58.14 & +8.70 \\
\midrule
\multirow{2}{*}{\href{https://huggingface.co/Qwen/Qwen3-Reranker-0.6B}{Qwen3-Reranker-0.6B}}
& \multirow{2}{*}{0.6B}
& Original 
& 44.21 & 43.85 & 45.62 & 40.55 & 43.56 & -- 
& 40.06 & 38.22 & 39.14 & -- \\
&
& Ours 
& 46.52 & 46.00 & 48.31 & 45.88 & 46.68 & +3.12
& 47.66 & 47.41 & 47.54 & +8.40 \\
\midrule
\multirow{2}{*}{\href{https://huggingface.co/BAAI/bge-reranker-v2-gemma}{bge-reranker-v2-gemma}}
& \multirow{2}{*}{3B}
& Original 
& 54.63 & 49.88 & 50.23 & 49.41 & 51.04 & --
& 51.17 & 54.31 & 52.74 & -- \\
&
& Ours
& 57.41 & 53.60 & 54.61 & 52.94 & 54.64 & +3.60
& 55.85 & 58.05 & 56.95 & +4.21 \\
\midrule
\multirow{2}{*}{\href{https://huggingface.co/Qwen/Qwen3-Reranker-4B}{Qwen3-Reranker-4B}}
& \multirow{2}{*}{4B}
& Original 
& 45.60 & 48.49 & 44.01 & 44.47 & 45.64 & -- 
& 46.49 & 49.71 & 48.10 & -- \\
&
& Ours 
& 58.80 & 57.54 & 57.14 & 59.06 & 58.14 & +12.50
& 69.30 & 66.67 & 67.99 & +19.89 \\
\midrule
\multirow{2}{*}{\href{https://huggingface.co/Qwen/Qwen3-Reranker-8B}{Qwen3-Reranker-8B}}
& \multirow{2}{*}{8B}
& Original 
& 49.54 & 44.32 & 42.17 & 42.59 & 44.66 & --
& 54.97 & 47.70 & 51.34 & -- \\
&
& Ours 
& \textbf{67.59} & \textbf{63.11} & \textbf{60.37} & \textbf{61.88} & \textbf{63.24} & \textbf{+18.58}
& \textbf{75.73} & \textbf{70.98} & \textbf{73.36} & \textbf{+22.02} \\
\bottomrule
\end{tabular}
}
\caption{Effect of mention anchoring across different rerankers. All models rerank the same top-$k$ candidate sets. For each reranker, the mention-anchored input explicitly marks the target mention span in the query. Results are R@1 (\%). Avg. is computed separately for each benchmark, and Gain denotes the improvement of mention-anchored input over the original input.}
\label{tab:ranker_anchor_transfer}
\end{table*}

\paragraph{Transfer across ranking interfaces.}
The preceding experiments use the pointwise interface of
Jina-Reranker-v3. Because this model also provides a native listwise
interface, we further examine whether mention anchoring transfers across
ranking formulations. As shown in Table~\ref{tab:jina_pointwise_listwise},
mention anchoring improves both pointwise and listwise reranking on EMEA and
Patent. The gains under listwise inference are smaller but remain consistent,
showing that our adaptation is not tied to independently scoring each
mention--candidate pair.
\begin{table*}[h]
\centering
\small
\setlength{\tabcolsep}{3.5pt}
\resizebox{\textwidth}{!}{
\begin{tabular}{llllcccccc cccc}
\toprule
\textbf{Dataset}$\rightarrow$ & & &
& \multicolumn{6}{c}{\textbf{EMEA}} 
& \multicolumn{4}{c}{\textbf{Patent}} \\
\cmidrule(lr){5-10}
\cmidrule(lr){11-14}
\textbf{Reranker}$\downarrow$ 
& \textbf{Size} 
& \textbf{Scoring}
& \textbf{Input}$\downarrow$ 
& \texttt{ES} & \texttt{FR} & \texttt{NL} & \texttt{DE} 
& \textbf{Avg.} & \textbf{Gain}
& \texttt{FR} & \texttt{DE} & \textbf{Avg.} & \textbf{Gain} \\

\midrule

\multirow{4}{*}{\href{https://huggingface.co/jinaai/jina-reranker-v3}{jina-reranker-v3}}
& \multirow{4}{*}{0.6B}
& \multirow{2}{*}{Pointwise}
& Original 
& 47.92 & 47.33 & 45.62 & 44.94 & 46.45 & -- 
& 51.46 & 47.41 & 49.44 & -- \\
&
&
& Ours
& 55.56 & 54.76 & 53.23 & 50.59 & 53.54 & \textbf{+7.09} 
& 60.53 & 55.75 & 58.14 & \textbf{+8.70}  \\

\cmidrule(lr){3-14}

&
&
\multirow{2}{*}{Listwise}
& Original 
& 48.15 & 49.19 & 47.24 & 45.18 & 47.44 & -- 
& 54.68 & 52.59 & 53.64 & -- \\
&
&
& Ours
& 53.94 & 54.99 & 53.69 & 46.82 & 52.36 & \textbf{+4.92} 
& 57.02 & 58.91 & 57.97 & \textbf{+4.33}  \\

\bottomrule
\end{tabular}
}
\caption{
Effect of mention-anchored adaptation under the pointwise and native listwise
interfaces of Jina-Reranker-v3. Both settings rerank the same top-$k$
candidate sets. Results are R@1 (\%). Avg. is computed separately for each
benchmark, and Gain denotes the improvement over the corresponding original
input.
}
\label{tab:jina_pointwise_listwise}
\end{table*}

\subsection{Exploratory Transfer to General-Purpose LLMs}
\label{app:general-purpose-transfer}

Although our main method and experiments focus on ranking-specialized
rerankers, we additionally examine whether mention anchoring can be applied
to general-purpose instruction-tuned LLMs. We prompt Llama-3.1-8B-Instruct\footnote{Llama-3.1-8B-Instruct: \url{https://huggingface.co/meta-llama/Llama-3.1-8B-Instruct}}
and Qwen3-8B\footnote{Qwen3-8B: \url{https://huggingface.co/Qwen/Qwen3-8B}} to assess retrieved candidates and compare them with
size-comparable \textsc{Qwen}3 rerankers. All models receive the same
retrieved candidate sets.
\begin{table*}[h]
\centering
\small
\setlength{\tabcolsep}{3.2pt}
\resizebox{\textwidth}{!}{
\begin{tabular}{llllcccccc cccc}
\toprule
\textbf{Dataset}$\rightarrow$ & & & 
& \multicolumn{6}{c}{\textbf{EMEA}} 
& \multicolumn{4}{c}{\textbf{Patent}} \\
\cmidrule(lr){5-10}
\cmidrule(lr){11-14}

\textbf{Model}$\downarrow$ 
& \textbf{Size}
& \textbf{Model type}
& \textbf{Input}$\downarrow$ 
& \texttt{ES} & \texttt{FR} & \texttt{NL} & \texttt{DE} 
& \textbf{Avg.} & \textbf{Gain}
& \texttt{FR} & \texttt{DE} & \textbf{Avg.} & \textbf{Gain} \\

\midrule

\multirow{2}{*}{
  \href{https://huggingface.co/Qwen/Qwen3-Reranker-4B}
  {Qwen3-Reranker-4B}
}
& \multirow{2}{*}{4B}
& \multirow{2}{*}{Ranker}
& Original 
& 45.60 & 48.49 & 44.01 & 44.47 & 45.64 & -- 
& 46.49 & 49.71 & 48.10 & -- \\
&
&
& Ours 
& 58.80 & 57.54 & 57.14 & 59.06 & 58.14 & \textbf{+12.50} 
& 69.30 & 66.67 & 67.99 & \textbf{+19.89} \\

\midrule

\multirow{2}{*}{
  \href{https://huggingface.co/meta-llama/Llama-3.1-8B-Instruct}
  {Llama-3.1-8B-Instruct}
}
& \multirow{2}{*}{8B}
& \multirow{2}{*}{General-purpose}
& Original 
& 26.85 & 25.29 & 27.65 & 25.65 & 26.36 & -- 
& 27.19 & 24.14 & 25.67 & -- \\
&
&
& Ours
& 45.60 & 42.69 & 46.54 & 44.71 & 44.89 & \textbf{+18.53} 
& 45.61 & 50.00 & 47.80 & \textbf{+22.13} \\

\midrule

\multirow{2}{*}{
  \href{https://huggingface.co/Qwen/Qwen3-Reranker-8B}
  {Qwen3-Reranker-8B}
}
& \multirow{2}{*}{8B}
& \multirow{2}{*}{Ranker}
& Original 
& 49.54 & 44.32 & 42.17 & 42.59 & 44.66 & --
& 54.97 & 47.70 & 51.34 & -- \\
&
&
& Ours 
& 67.59 & 63.11 & 60.37 & 61.88 & 63.24 & \textbf{+18.58}
& 75.73 & 70.98 & 73.36 & \textbf{+22.02} \\

\midrule

\multirow{2}{*}{
  \href{https://huggingface.co/Qwen/Qwen3-8B}
  {Qwen3-8B}
}
& \multirow{2}{*}{8B}
& \multirow{2}{*}{General-purpose}
& Original 
& 38.89 & 42.69 & 41.24 & 41.88 & 41.18 & --
& 50.88 & 52.87 & 51.88 & -- \\
&
&
& Ours 
& 78.81 & 78.89 & 76.04 & 77.65 & 77.85 & \textbf{+36.67} 
& 79.53 & 77.10 & 78.32 & \textbf{+26.44} \\

\bottomrule
\end{tabular}
}
\caption{
Exploratory evaluation of mention-anchored adaptation with general-purpose
LLMs and ranking-specialized rerankers. All models receive the same top-$k$
candidate sets. Results are R@1 (\%). Avg. is computed separately for each
benchmark, and Gain denotes the improvement over the original input.
General-purpose LLM results should be interpreted with caution because these
models may assign identical highest scores to multiple candidates.
}
\label{tab:general_ranker_anchor_transfer}
\end{table*}

Table~\ref{tab:general_ranker_anchor_transfer} shows that mention anchoring
substantially improves both general-purpose models. This suggests that
explicitly identifying the target mention is also useful beyond
ranking-specialized architectures. However, general-purpose LLMs frequently
assign the same highest score to multiple candidates, making top-1 prediction
sensitive to tie breaking. We therefore treat these results as exploratory
and use ranking-specialized rerankers in our main experiments.

\subsection{Mention Marker Ablation}
\label{app:maptag}

Having shown that mention anchoring improves reranking across different rerankers, we further examine which marker should be used to indicate the target span. Table~\ref{tab:marker_ablation_emea_patent} compares several mention-marking strategies while keeping the retriever, candidate set, and reranker fixed. All explicit markers improve over the no-marker setting, confirming that directing the reranker to the target span is important for the entity linking task. The \texttt{<tgt>}...\texttt{</tgt>} marker performs best across all EMEA and Patent settings. We attribute this gain to the fact that \texttt{<tgt>}...\texttt{</tgt>} follows a familiar span-highlighting convention (similar to HTML-style delimiters), making it easier for the reranker to reliably focus on the target mention when comparing candidates.
\begin{table*}[t]
\centering
\small
\setlength{\tabcolsep}{4pt}
\begin{tabular}{lcccccc}
\toprule
\textbf{Mention marker} 
& \multicolumn{4}{c}{\textbf{EMEA}} 
& \multicolumn{2}{c}{\textbf{Patent}} \\
\cmidrule(lr){2-5}
\cmidrule(lr){6-7}
& \texttt{ES} & \texttt{FR} & \texttt{NL} & \texttt{DE}
& \texttt{FR} & \texttt{DE} \\
\midrule
No explicit marker & 62.70 & 58.47 & 53.23 & 56.71 & 67.54 & 60.92 \\
\texttt{*...*} & 63.89 & 61.95 & 57.83 & 59.76 & 71.93 & 64.37 \\
\texttt{<mention>}...\texttt{</mention>} & 65.97 & 62.18 & 58.99 & 60.12 & 73.39 & 68.97 \\
\texttt{<tgt>}...\texttt{</tgt>} & \textbf{67.59} & \textbf{63.11} & \textbf{60.37} & \textbf{61.88} & \textbf{75.73} & \textbf{70.98} \\
\bottomrule
\end{tabular}
\caption{Effect of different mention markers for reranking. All settings rerank the same top-$k$ candidate sets and differ only in how the target mention is marked in the query. Results are R@1 (\%). Test reranker is \textsc{Qwen}3-Reranker.}

\label{tab:marker_ablation_emea_patent}
\end{table*}

\section{Additional Experimental Results}
\label{app:result_details}

\subsection{Standard Deviations over Random Seeds}
\label{app:rs}

For \ModelName{}, the main tables in Section~\ref{sec:mr} report mean R@1 over three random seeds. Table~\ref{tab:seed_std_all} provides the corresponding mean and standard deviation values.  For \ModelName{} Retriever, we train the retriever with three different random seeds and evaluate R@1 for each dataset/language in each run. We then compute the mean and standard deviation of the three R@1 values for each reported column. For \ModelName{}, the reranker is fixed and not trained by us. We apply the same fixed reranker to the candidate sets produced by the three retriever runs, evaluate the resulting end-to-end R@1 for each run, and compute the mean and standard deviation across the three runs. All values are reported as mean (standard deviation) in R@1 percentage points.

\begin{table}[h]
\centering
\small
\setlength{\tabcolsep}{4pt}
\begin{tabular}{llcc}
\toprule
\textbf{Dataset} & \textbf{Lang.} & \textbf{\ModelName{} Retriever} & \textbf{\ModelName{}} \\
\midrule
XL-BEL & \texttt{EN} & 89.9 (0.42) & 91.2 (0.23) \\
XL-BEL & \texttt{ES} & 63.7 (0.36) & 69.6 (0.25) \\
XL-BEL & \texttt{DE} & 45.4 (0.40) & 53.0 (1.25) \\
XL-BEL & \texttt{FI} & 31.0 (0.38) & 35.9 (0.61) \\
XL-BEL & \texttt{RU} & 49.5 (0.57) & 56.6 (0.46) \\
XL-BEL & \texttt{TR} & 54.5 (0.57) & 60.3 (0.85) \\
XL-BEL & \texttt{KO} & 35.4 (0.37) & 39.6 (0.35) \\
XL-BEL & \texttt{ZH} & 35.0 (0.49) & 44.5 (0.38) \\
XL-BEL & \texttt{JA} & 32.6 (0.40) & 39.1 (0.46) \\
XL-BEL & \texttt{TH} & 41.2 (0.85) & 48.6 (1.10) \\
\midrule
EMEA & \texttt{ES} & 63.3 (0.13) & 68.9 (0.13) \\
EMEA & \texttt{FR} & 60.6 (0.13) & 65.4 (1.16) \\
EMEA & \texttt{NL} & 60.0 (1.27) & 65.6 (0.74) \\
EMEA & \texttt{DE} & 60.2 (2.46) & 64.6 (0.49) \\
\midrule
Patent & \texttt{FR} & 73.0 (0.17) & 78.8 (0.34) \\
Patent & \texttt{DE} & 67.2 (0.00) & 74.2 (0.72) \\
\midrule
WikiMed-DE & \texttt{DE} & 63.9 (1.02) & 67.0 (0.21) \\
WikiMed-DE & \texttt{DE-wiki} & 78.1 (0.12) & 79.8 (0.30) \\
\midrule
MedMentions & \texttt{EN} & 53.3 (0.03) & 60.1 (0.47) \\
\bottomrule
\end{tabular}
\caption{R@1 (\%) over three random seeds. Each value is reported as mean (standard deviation). \textbf{Lang.} denotes language. For \textbf{\ModelName{} Retriever}, the standard deviation is computed over three independently trained retrievers. For \textbf{\ModelName{}}, the reranker is fixed; the standard deviation is computed after applying the same reranker to the candidate sets produced by the three retriever runs.}
\label{tab:seed_std_all}
\end{table}

\subsection{Results with the Case-Preserving UMLS 2020AA KB}
\label{app:cased_kb}
Table~\ref{tab:emaa_patent_wiki_cased} repeats the comparison of Table~\ref{tab:emaa_patent_wiki} on the case-preserving version of UMLS 2020AA (Appendix~\ref{experiment_setup}). With the case-preserving KB, the \ModelName{} retriever improves over the lowercased KB on EMEA (62.4 vs.\ 61.1 average R@1) and WikiMed-DE (64.4 vs.\ 63.9), while it is slightly lower on Patent (69.5 vs.\ 70.1). A likely reason is that the Wikidata-derived aliases used to train the retriever keep their original casing and the XLM-R tokenizer is case-sensitive, so the retriever learns cased surface forms that match the case-preserving KB more closely. \ModelName{} with reranking remains the best system in every column under this KB as well. Since mentions keep their original casing, the reproduced baselines also benefit from the case-preserving KB.

\begin{table*}[t]
\centering
\small
\setlength{\tabcolsep}{3.2pt}

\begin{tabular*}{\textwidth}{@{\extracolsep{\fill}}
l *{4}{S[table-format=2.1]} S[table-format=2.1] *{2}{S[table-format=2.1]} S[table-format=2.1] S[table-format=2.1]
}
\toprule
\textbf{Dataset}$\rightarrow$
& \multicolumn{4}{c}{\textbf{EMEA}}
& \textbf{Avg}
& \multicolumn{2}{c}{\textbf{Patent}}
& \textbf{Avg}
& \textbf{WikiMed-DE} \\
\cmidrule(lr){2-5}\cmidrule(lr){7-8}\cmidrule(lr){10-10}
\textbf{Method}$\downarrow$
& \texttt{ES} & \texttt{FR} & \texttt{NL} & \texttt{DE}
&
& \texttt{FR} & \texttt{DE}
&
& \texttt{DE} \\
\midrule
BERGAMOT~\cite{sakhovskiy-etal-2024-biomedical}
& \underline{65.7} & 59.9 & 59.0 & 63.5
& 62.0
& 69.3 & 66.7
& 68.0
& 59.3 \\
SapBERT$_{\mbox{de}}$~\cite{mustafa-etal-2024-leveraging}
& 65.0 & 60.3 & \underline{60.6} & \underline{64.9}
& \underline{62.7}
& 67.0 & 66.7
& 66.8
& 61.0 \\
\midrule
\ModelName{} retriever (\textbf{ours})
& 63.9 & \underline{61.6} & 60.1 & 63.8
& 62.4
& \underline{70.9} & \underline{68.2}
& \underline{69.5}
& \underline{64.4} \\
\ModelName{} retriever+reranker (\textbf{ours})
& \textbf{66.0} & \textbf{63.9} & \textbf{63.7} & \textbf{66.2}
& \textbf{64.9}
& \textbf{77.0} & \textbf{73.5}
& \textbf{75.2}
& \textbf{67.6} \\
\bottomrule
\end{tabular*}

\caption{R@1 (\%) on EMEA, Patent and WikiMed-DE with the case-preserving version of UMLS 2020AA (Appendix~\ref{experiment_setup}); Table~\ref{tab:emaa_patent_wiki} reports the same benchmarks with the lowercased version. All numbers are obtained by us under the same evaluation protocol. The best results are bold, second best are underlined. For \ModelName{}, we report the mean over three random seeds.}
\label{tab:emaa_patent_wiki_cased}
\end{table*}

\subsection{Additional English BEL Evaluation on MedMentions}
\label{app:medmentions}
\begin{table}[h]
\centering
\small
\setlength{\tabcolsep}{5pt}
\resizebox{\columnwidth}{!}{
\begin{tabular}{l S[table-format=2.1]}
\toprule

\textbf{Method}$\downarrow$ & {MedMentions (EN)} \\
\midrule
\SapBERTmulti{}~\cite{DBLP:conf/acl/0001VKC20} & 50.3$^{\divideontimes}$  \\
GenBioEL~\cite{DBLP:journals/jbi/YuanZSLWY22}
& 70.7 \\
ArboEL~\cite{DBLP:conf/naacl/0003AMM22}
& \textbf{74.7} \\
Prompt-BioEL~\cite{DBLP:conf/aaai/XuCH23}
&72.6 \\
BioPro~\cite{DBLP:journals/taslp/ZhuQFCHX24}
& 66.5 \\
GenBioEL+ANGEL~\cite{DBLP:conf/acl/KimKPLSK25}
& \underline{73.3} \\
\midrule
\ModelName{} retriever (\textbf{ours})
& 53.3 \\
\ModelName{} retriever+reranker (\textbf{ours})
& 60.1\\
\bottomrule
\end{tabular}
}
\caption{R@1 (\%) on MedMentions. $\divideontimes$ indicates reproduced results. Best result is bold, second best is underlined. All listed methods are trained on MedMentions except for ours and \SapBERTmulti{}. For \ModelName{}, we report the mean over three random seeds; standard deviations are provided in Appendix~\ref{app:rs}.}
\label{tab:mm}
\end{table}
Table~\ref{tab:mm} reports R@1 on MedMentions, which we use as an English BEL transfer check. Unlike the fully supervised MedMentions baselines, \ModelName{} does not use the MedMentions training set. In this annotation-free setting, our \ModelName{} retriever improves over \SapBERTmulti{} from 50.3 to 53.3 (+3.0), and the full retrieve--rerank system reaches 60.1. Fully supervised systems remain stronger because they are trained end-to-end on MedMentions annotations, so this comparison is not supervision-matched. Nevertheless, the gains over \SapBERTmulti{} indicate that Wikidata-derived alias supervision and mention-anchored reranking also benefit English BEL without task-specific MedMentions training.

\subsection{Translate Baseline}
\label{app:translation}
Biomedical information is produced across many languages, whereas biomedical knowledge bases and many existing BEL systems remain largely English-centric.
Cross-lingual BEL aims to bridge this gap by directly linking non-English biomedical mentions to standardized concepts. An alternative is to first translate non-English mentions into English and then apply an English BEL
system. However, translation introduces an additional processing step and may alter specialized terminology, abbreviations, or language-specific lexical forms. We therefore evaluate this alternative on XL-BEL.

\paragraph{Experimental Setup.}
We translate non-English XL-BEL mentions into English using
ArgosTranslate\footnote{\url{https://github.com/argosopentech/argos-translate}} and evaluate two English BEL systems with different supervision settings.
SapBERT$_{\mathrm{en}}$~\citep{DBLP:conf/naacl/LiuSMBC21} is a benchmark-agnostic baseline trained on English biomedical aliases in UMLS and does not require training annotations from the target benchmark. We also evaluate
ANGEL~\citep{DBLP:conf/acl/KimKPLSK25}, a more recent strong English BEL system. Unlike SapBERT$_{\mathrm{en}}$ and \ModelName{}, ANGEL is designed
with benchmark-specific supervised training. Here, we evaluate its ability to transfer to XL-BEL without training on XL-BEL annotations. We use the ANGEL checkpoint trained on BC5CDR. In comparison,
\ModelName{} directly processes the original multilingual mentions without translation.

\begin{table*}[t]
\centering
\small
\setlength{\tabcolsep}{3.0pt}

\begin{tabular*}{\textwidth}{@{\extracolsep{\fill}}
ll
*{10}{S[table-format=2.1]}
S[table-format=2.1]
}
\toprule
\textbf{Dataset}$\rightarrow$
&
& \multicolumn{10}{c}{\textbf{XL-BEL}}
& \textbf{Avg.} \\
\cmidrule(lr){3-12}

\textbf{Model}$\downarrow$
& \textbf{Input}
& \texttt{EN}
& \texttt{ES}
& \texttt{DE}
& \texttt{FI}
& \texttt{RU}
& \texttt{TR}
& \texttt{KO}
& \texttt{ZH}
& \texttt{JA}
& \texttt{TH}
& \\
\midrule

SapBERT$_{\mathrm{en}}$~\citep{DBLP:conf/naacl/LiuSMBC21}

& Translated
& 78.7 & 34.0 & 40.7 & 19.1 & 29.7
& 19.8 & 35.8 & 89.2 & 8.7 & 24.7
& 38.0 \\


\midrule

ANGEL~\citep{DBLP:conf/acl/KimKPLSK25}

& Translated
& 13.1 & 3.0 & 2.0 & 1.0 & 4.0
& 3.0 & 1.0 & 1.0 & 1.0 & 2.0
& 3.1 \\


\midrule

\ModelName{} (\textbf{ours})
& Original
& 89.9
& 63.7
& 45.4
& 31.0
& 49.5
& 54.5
& 35.4
& 35.0
& 32.6
& 41.2
& 47.8 \\

\bottomrule
\end{tabular*}

\caption{
R@1 (\%) on XL-BEL using English translation baselines and direct cross-lingual linking. For SapBERT$_{\mathrm{en}}$ and ANGEL, non-English
mentions are translated into English with ArgosTranslate before entity linking.
\ModelName{} directly processes the original multilingual mentions. ANGEL is
evaluated without XL-BEL-specific supervised training.
}
\label{tab:xlbel_translation}
\end{table*}

\paragraph{Results.}
As shown in Table~\ref{tab:xlbel_translation}, SapBERT$_{\mathrm{en}}$ with translated inputs achieves an average R@1 of 38.0, compared with 47.8 for \ModelName{} on the original multilingual mentions. The effectiveness of translation varies substantially across languages, indicating that downstream linking performance is highly sensitive to translation quality.

\begin{figure}[t]
    \centering
    \includegraphics[width=\columnwidth]{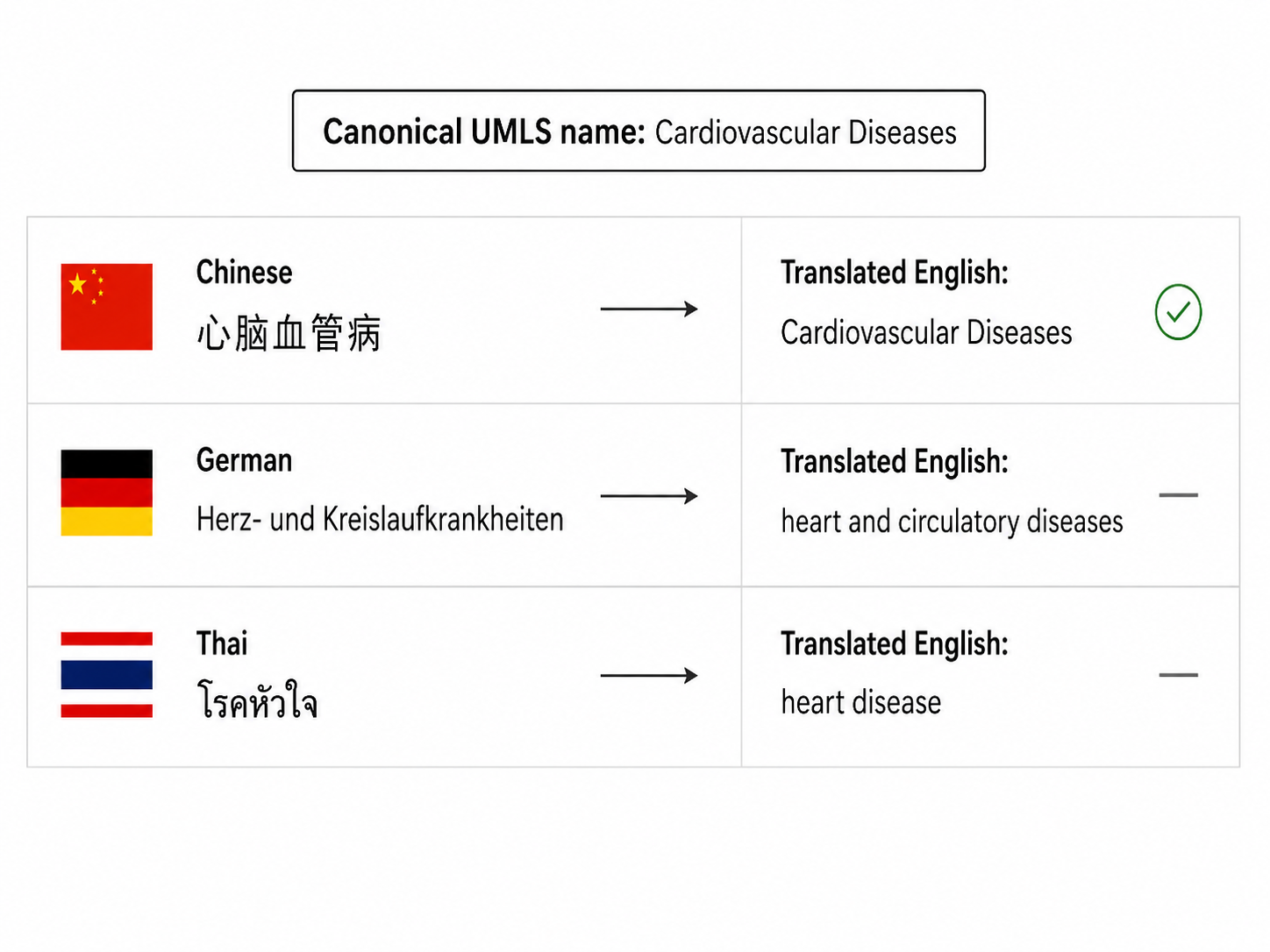}
    \caption{
    Examples of how translation quality can affect biomedical entity linking.
    All three mentions refer to the same UMLS concept, whose canonical English
    name is ``Cardiovascular Diseases.'' The Chinese mention is translated
    directly to the canonical term, whereas the German and Thai mentions are
    translated into more descriptive or general expressions. Such translation
    differences can affect downstream linking performance.
    }
    \label{fig:translation_examples}
\end{figure}

Biomedical mentions are particularly challenging to translate because they often contain specialized terminology, abbreviations, and language-specific lexical conventions. Translation errors can alter the intended concept, while even semantically reasonable translations may produce descriptive or more general expressions that differ from canonical English aliases. Figure~\ref{fig:translation_examples} illustrates this issue. The Chinese mention is translated directly to ``Cardiovascular Diseases,'' matching the canonical UMLS name. In contrast, the German and Thai mentions are translated into broader or more descriptive expressions.

This pattern is also reflected in the cross-language results. For most languages, linking translated mentions with SapBERT$_{\mathrm{en}}$ performs substantially below direct cross-lingual linking with \ModelName{}. Chinese stands out as a clear exception, where translation raises R@1 to 89.2. Many Chinese biomedical mentions in XL-BEL are compositional terms whose translations closely resemble canonical English biomedical terminology, making them easier to match to English KB aliases. For other languages, translation more often introduces lexical variation, overly general expressions, or semantic errors, which can make downstream entity linking more difficult.

ANGEL achieves substantially lower performance when transferred directly to
XL-BEL without XL-BEL-specific training. Unlike the benchmark-agnostic
SapBERT$_{\mathrm{en}}$, ANGEL is optimized with supervised data from its
training benchmark, BC5CDR. XL-BEL is an evaluation-only benchmark and provides
no task-specific training data, making such supervised adaptation unavailable.
This highlights a practical limitation of fully supervised English BEL systems:
many cross-lingual benchmarks and real-world low-resource settings do not provide
sufficient annotated data for task-specific training. In contrast, \ModelName{}
is trained without benchmark-specific supervision and can be directly applied
across languages and benchmarks.

Taken together, these results show that English translation can be effective for
some languages, but does not consistently replace direct cross-lingual BEL.
They also highlight the advantage of benchmark-agnostic cross-lingual modeling
in settings where target-language annotations are limited or unavailable.

\section{Analysis of Retrieval Improvements}
\label{app:error_analysis}
\subsection{Do the retrieval gains come from cross-lingual alias bridges?}
To better understand why the \ModelName{} retriever improves over \SapBERTmulti{}, we analyze the 1,416 improved mentions on XL-BEL, where our retriever yields the largest gains. We define an \emph{improved mention} as a test mention for which \SapBERTmulti{} fails to rank the gold CUI at top-1, while the \ModelName{} retriever ranks the gold CUI at top-1.

We group these improved mentions according to the Wikidata-derived alias coverage of their gold CUI after decontamination. Group A contains mentions whose gold CUI has only same-language Wikidata aliases. Group B contains mentions whose gold CUI has aliases in at least one other language, providing potential cross-lingual alignment bridges. Group C contains mentions whose gold CUI has no Wikidata-derived alias coverage. Table~\ref{tab:improved_mentions_group_composition} reports the group composition for each language.

The results show that the improvements are not driven by same-language alias coverage alone: Group A is empty for all languages. Instead, 82.3\% of improved mentions belong to Group B, where the gold CUI is connected to aliases in other languages. This supports our cross-lingual alias bridge interpretation: multilingual aliases can anchor the correct concept across languages and help unseen non-English mentions retrieve the correct CUI, even when direct same-language coverage is limited. The remaining 17.7\% of improvements fall into Group C, suggesting that continued training also improves the global embedding space beyond concepts directly covered by Wikidata aliases, although most improved mentions are associated with cross-lingual alias bridges.
\begin{table}[!t]
\centering
\small
\setlength{\tabcolsep}{3pt}
\begin{tabular}{lrrrr}
\toprule
\textbf{Lang.} & \textbf{Imp. (\%)} & 
\textbf{Group A} & 
\textbf{Group B} & 
\textbf{Group C} \\
\midrule
\texttt{DE} & 152 (15.2) & 0 & 127 (83.6) & 25 (16.4) \\
\texttt{EN} & 68  (6.8)  & 0 & 59  (86.8) & 9  (13.2) \\
\texttt{ES} & 74  (7.4)  & 0 & 57  (77.0) & 17 (23.0) \\
\texttt{FI} & 143 (14.3) & 0 & 111 (77.6) & 32 (22.4) \\
\texttt{JA} & 118 (11.8) & 0 & 99  (83.9) & 19 (16.1) \\
\texttt{KO} & 197 (19.7) & 0 & 150 (76.1) & 47 (23.9) \\
\texttt{RU} & 152 (15.2) & 0 & 120 (78.9) & 32 (21.1) \\
\texttt{TH} & 204 (20.4) & 0 & 186 (91.2) & 18 (8.8) \\
\texttt{TR} & 121 (12.1) & 0 & 107 (88.4) & 14 (11.6) \\
\texttt{ZH} & 187 (18.7) & 0 & 150 (80.2) & 37 (19.8) \\
\midrule
\textbf{All} & 1416 (14.2) & 0 & 1166 (82.3) & 250 (17.7) \\
\bottomrule
\end{tabular}
\caption{Group composition of improved mentions on XL-BEL. \textbf{Lang.} denotes the language code. \textbf{Imp. (\%)} denotes the number of improved mentions and their percentage among test mentions in that language. Group A: only same-language Wikidata coverage; Group B: Wikidata coverage in at least one other language; Group C: no Wikidata coverage. Percentages in each group are computed over improved mentions.}
\label{tab:improved_mentions_group_composition}
\end{table}

We extend the analysis to the full XL-BEL dataset. Table~\ref{tab:other_language_alias_bins} groups all 10,000 XL-BEL mentions by the Wikidata-derived alias coverage of their gold CUIs. The \textit{No coverage} bin contains mentions whose gold CUI has no Wikidata-derived aliases in the retriever training data. The \textit{Same-lang only} bin contains mentions whose gold CUI has aliases only in the mention language. The remaining bins correspond to gold CUIs with increasing amounts of other-language alias coverage.

BioELX improves over \SapBERTmulti{} in every non-empty coverage bin. The gain in the \textit{No coverage} bin (+9.3 R@1) suggests that Wikidata-derived training does not only benefit CUIs directly covered by the added aliases. Instead, continued contrastive training may also improve the global geometry of the embedding space, reducing the model's reliance on superficial lexical similarity and improving generalization to uncovered concepts. At the same time, the largest gain appears in the 21+ bin (+12.3 R@1), where gold CUIs have broad other-language alias coverage. This supports the cross-lingual bridge interpretation: richer multilingual alias bridges provide stronger concept-level supervision. Together with Table~\ref{tab:improved_mentions_group_composition}, these results suggest that the retrieval gains are not explained by same-language alias memorization, but are largely associated with cross-lingual alias bridges and improved multilingual representation learning.

\begin{table*}[!t]
\centering
\small
\setlength{\tabcolsep}{5pt}
\begin{tabular}{lrrrrr}
\toprule
\textbf{Alias coverage} & 
\textbf{Mentions} & 
\textbf{\% test mentions} & 
\textbf{\SapBERTmulti{} R@1} & 
\textbf{BioELX R@1} & 
\textbf{$\Delta$} \\
\midrule
No coverage     & 2,263 & 22.6 & 30.4 & 39.7 & +9.3 \\
Same-lang only  & 0 & 0.0 & -- & -- & -- \\
2--5 other langs  & 182   & 1.8  & 80.8 & 84.1 & +3.3 \\
6--20 other langs & 949   & 9.5  & 51.7 & 61.0 & +9.3 \\
21+ other langs   & 6,606 & 66.1 & 35.7 & 48.0 & +12.3 \\
\bottomrule
\end{tabular}
\caption{XL-BEL mentions grouped by the Wikidata-derived alias coverage of the gold CUI in the retriever training data. \textbf{\% test mentions} denotes the percentage of XL-BEL test mentions in each coverage bin. \textit{No coverage} means that the gold CUI has no Wikidata-derived aliases in the training data. \textit{Same-lang only} means that aliases are available only in the mention language. The remaining bins indicate the number of languages different from the mention language in which the gold CUI has aliases. Results are R@1 (\%). $\Delta$ denotes the absolute improvement of BioELX over SapBERT.}
\label{tab:other_language_alias_bins}
\end{table*}

\subsection{Dataset Ambiguity and Retriever Gains}
\label{app:retriever_gain_dataset}

To understand why the retriever gains are largest on XL-BEL, we compare the degree of surface-form ambiguity across benchmarks. We define a mention as ambiguous if the same mention string appears in the benchmark with more than one gold CUI. This statistic measures cases where the mention string alone does not uniquely determine the target entity, and where context is more likely to be needed for disambiguation.

\begin{table}[H]
\centering
\small
\setlength{\tabcolsep}{4pt}
\begin{tabular}{lrrr}
\toprule
\textbf{Dataset} & \textbf{Mentions} & \textbf{Ambiguous} & \textbf{\%} \\
\midrule
XL-BEL & 10,000 & 69 & 0.7 \\
WikiMed-DE & 41,120 & 2,022 & 4.9 \\
Patent & 1,043 & 174 & 16.7 \\
EMEA & 2,155 & 665 & 30.9 \\
\bottomrule
\end{tabular}
\caption{Surface-form ambiguity across benchmarks. We count a mention as ambiguous if its mention string appears in the benchmark with more than one gold CUI.}
\label{tab:dataset_ambiguity}
\end{table}

Table~\ref{tab:dataset_ambiguity} shows that XL-BEL has the lowest surface-form ambiguity: only 0.7\% of its mentions share the same surface form with another entity linked to a different CUI. WikiMed-DE is also relatively low at 4.9\%. In contrast, EMEA and Patent are substantially more ambiguous, with 30.9\% and 16.7\% ambiguous mentions, respectively.

These differences help explain retrieval gains in the main experiments. Since our retriever is trained from aliases, it primarily improves alias-level matching. This is especially effective for benchmarks with low surface-form ambiguity, where the mention string itself provides strong evidence for the target entity. In more ambiguous benchmarks such as EMEA and Patent, the same mention string more often maps to different CUIs, so improving alias matching alone is less sufficient, and context-aware reranking becomes more important. This also helps explain why EMEA and Patent benefit more from context-aware reranking than from improvements in retrieval alone.

\section{Additional Ablations}
\label{app:ablation}
\subsection{How Does the Retriever Affect Reranking?}
\label{app:retre_affect}

Since our reranker only operates on the top-$k$ retrieved candidates, the retriever determines an upper bound on the achievable reranking accuracy: if the gold entity is not contained in $C_k(m)$, no reranking model can recover it. 
To quantify the dependency of the reranker on the retriever, we plug candidate sets produced by different retrievers into the same \textsc{Qwen}3-Reranker and compare the final R@1. 
We consider a range of retrievers with varying strengths, including BM25~\cite{robertson2009bm25} (sparse lexical retrieval), mBART (a weaker neural baseline), BERGAMOT (a strong cross-lingual system), and our \ModelName{} retriever.
\begin{table}[t]
\centering
\small
\setlength{\tabcolsep}{4pt}
\begin{tabular}{lccc}
\toprule
\textbf{Retriever} & \textbf{R@1} & \textbf{R@64} & \textbf{+ Reranker R@1} \\
\midrule
BM25 & 19.7 & 26.9 & 26.1 \\
mBART & 17.2 & 26.2 & 25.9 \\
BERGAMOT & 35.6 & 46.6 & 37.3 \\
\ModelName{} retriever & 50.1 & 70.7 & \textbf{54.8} \\
\bottomrule
\end{tabular}
\caption{Effect of retriever quality on reranking. The reranker is fixed (\textsc{Qwen}3-Reranker with the same prompt and $k{=}64$). R@$k$ provides an upper bound on reranked R@1. We report average results over the XL-BEL benchmark.}
\label{tab:ablation_retriever_rerank}
\end{table}

\noindent Table~\ref{tab:ablation_retriever_rerank} reports each retriever's R@1 and R@$k$ (with $k{=}64$) together with the reranked R@1 using identical reranking settings on XL-BEL. 
The results show a clear trend: stronger retrievers yield substantially better reranked performance, largely because they place the gold entity in the top-$k$ set more often. 
For example, mBART has relatively low recall at $k{=}64$ (26.2), whereas BERGAMOT improves this to 46.6, and our retriever further increases it to 70.7; correspondingly, reranking accuracy improves as candidate coverage increases.
These findings highlight that improving candidate retrieval is a prerequisite for effective contextual reranking in large biomedical knowledge bases.


\subsection{Does Entity-Side Metadata Help Reranking?}
\label{app:entity_info}
Except for entity aliases, the knowledge base also provides entity metadata, including type and description. We examine whether providing additional entity metadata can improve LLM reranking beyond using the entity alias alone. 
Specifically, we compare three input settings for the reranker: (i) entity alias only (A), (ii) alias plus semantic type (A+T), and (iii) alias plus type and textual description (A+T+D).
\begin{table}[H]
\centering
\small
\setlength{\tabcolsep}{6pt}
\begin{tabular}{l c}
\toprule
\textbf{Entity-side input to reranker} & \textbf{R@1} \\
\midrule
Alias only (A) & 54.8 \\
Alias + Semantic Type (A+T) & 53.5 \\
Alias + Type + Description (A+T+D) & 52.5 \\
\bottomrule
\end{tabular}
\caption{Effect of adding entity-side metadata to the \textsc{Qwen}3-Reranker on XL-BEL. All variants rerank the same top-$k$ candidate sets produced by the same retriever (fixed $k$).}
\label{tab:addst}
\end{table}
As shown in Table~\ref{tab:addst}, adding entity types or descriptions does not improve performance and even slightly degrades R@1. 
We attribute this to two factors. 
First, within the top-$k$ retrieved candidates, many entities already share the same (or highly similar) semantic types, so type information provides little discriminative signal but adds extra tokens that can distract the reranker. 
Second, entity descriptions have limited coverage for the entities appearing in our evaluation: in XL-BEL, only a small fraction of entities provide non-empty descriptions (roughly 10\%), making the A+T+D setting inconsistent across candidates and often uninformative.

\section{Use of AI Assistants}
We used AI assistants only for writing polishing, such as grammar correction and minor rephrasing. 

\section{Additional Figures}
\label{app:large_figures}

\begin{figure*}[ht]
  \includegraphics[width=\textwidth]{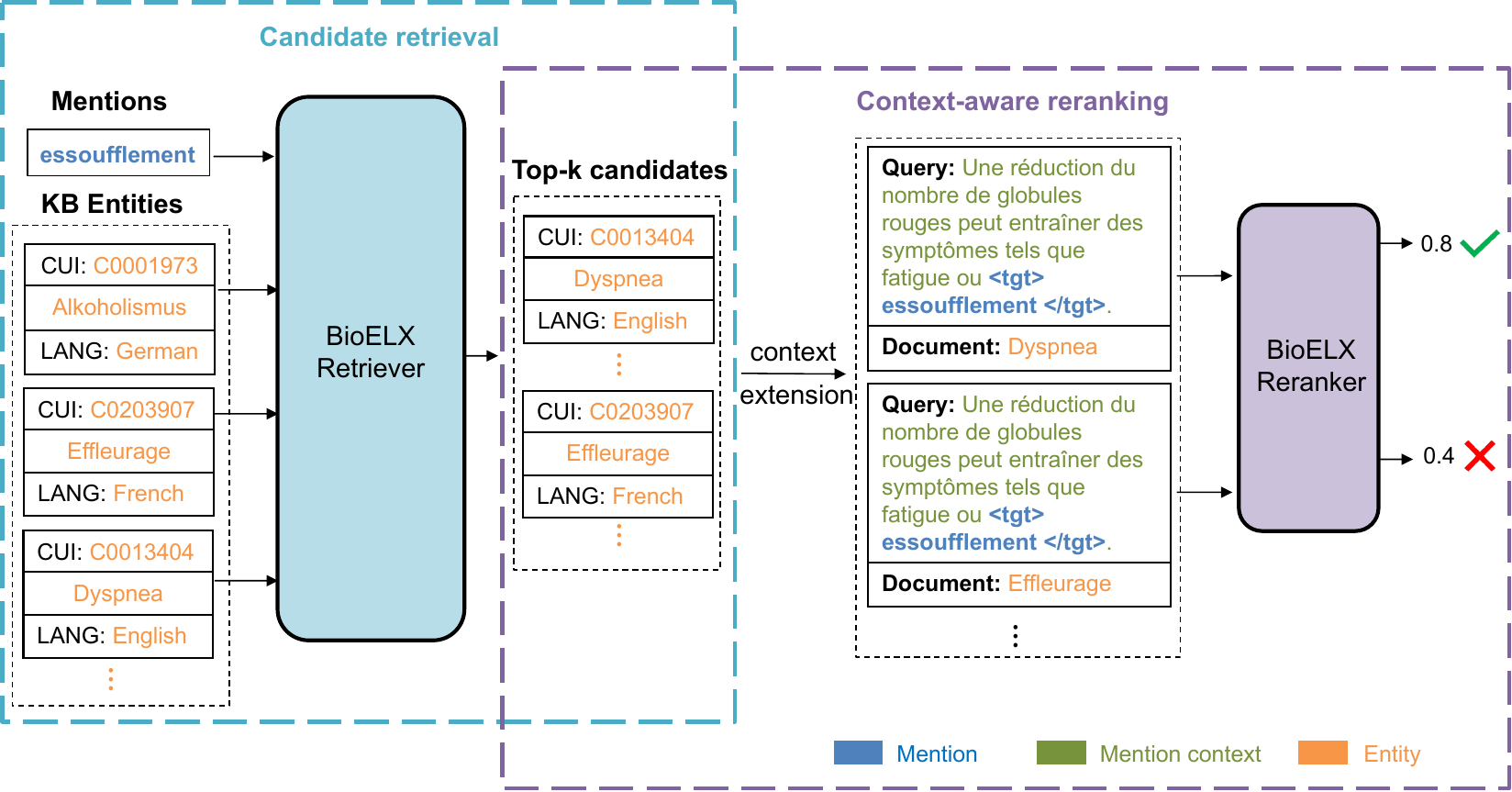}
  \caption{The two-stage \ModelName{} pipeline, illustrated for a single mention. The retriever embeds the mention and multilingual entity aliases into a shared space, and returns the top-$k$ aliases with associated entities by cosine similarity. For each candidate alias, the reranker scores a \textit{mention-anchored query} --- the mention and its surrounding context --- against a document built from that alias, and selects the entity associated with the highest-scoring alias. }
  \label{fig:pipeline_large}
\end{figure*}

\begin{figure*}[ht]
  \includegraphics[width=\textwidth]{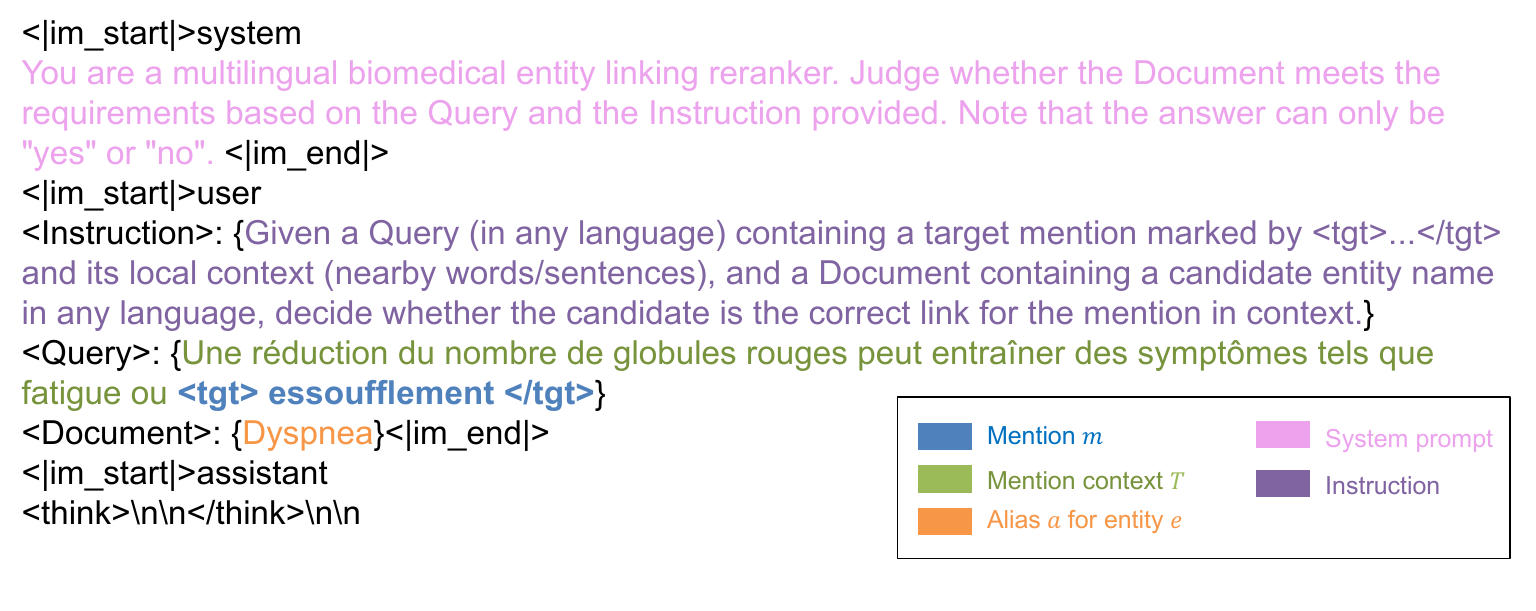}
  \caption{Prompt format for the \ModelName{} Reranker, instantiated with the \textsc{Qwen}3-Reranker template (black text). The \textbf{query} contains the biomedical text $\Text$ with the target mention $\Mention$ marked by \texttt{<tgt>}...\texttt{</tgt>}, and the \textbf{document} contains an alias $\Alias$ of the candidate entity $\Entity$, so the reranker scores each mention-candidate pair. Our ``document'' is intentionally short: many KB entities lack rich textual metadata, so we provide only the entity alias.}
  
  \label{fig:temp_large}
\end{figure*}

\begin{figure*}[ht]
  \includegraphics[width=\textwidth]{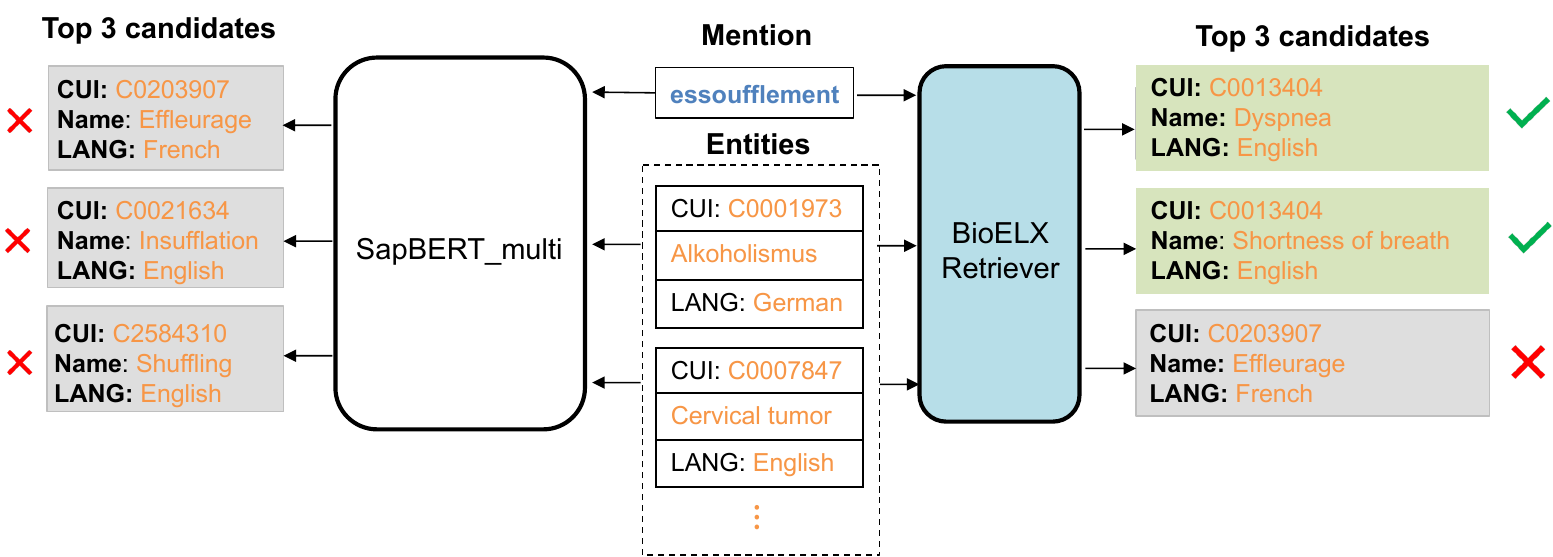}
  \caption{An example of retrieval. For the French mention \textit{essoufflement}, SapBERT retrieves top candidates based on surface similarity but misses the correct concept. Our retriever, enhanced with Wikidata-derived multilingual aliases, retrieves candidates that include the correct CUI (Dyspnea) among the top results.}
  \label{fig:sapfail_large}
\end{figure*}

\end{document}